\documentclass[letterpaper, 10 pt, conference]{ieeeconf}  

\IEEEoverridecommandlockouts                              

\usepackage{graphics} 
\usepackage{epsfig} 
\usepackage{times} 
\usepackage{amsmath} 

\usepackage{amssymb}  
\usepackage{bbm}
\usepackage[dvipsnames]{xcolor}
\usepackage[utf8]{inputenc}
\usepackage{subcaption}
\usepackage{pdfpages}
\usepackage{paralist}
\usepackage[font=small,labelfont=bf]{caption}
\usepackage{array}
\usepackage{booktabs}
\usepackage{lipsum}
\usepackage{caption}
\usepackage{siunitx}
\usepackage{cite}
\usepackage{mathtools}
\usepackage{xcolor}
\usepackage{colortbl}
\usepackage{blkarray, bigstrut}
\usepackage{svg}
\let\labelindent\relax
\usepackage{enumitem}
\usepackage[bookmarksopen, bookmarksdepth=2, breaklinks=true ]{hyperref}
\RequirePackage{color,graphicx}
\definecolor{linkcolour}{rgb}{0.2,0.285,0.918}
\hypersetup{colorlinks,linkcolor={blue},citecolor={linkcolour},urlcolor={blue}} 
\usepackage{graphicx}
\usepackage{multirow}

\newcommand{\DUTT}{\textcolor{black}}

\AtBeginEnvironment{quote}{\raggedright}

\usepackage{tcolorbox} 
\tcbuselibrary{skins, breakable} 
\newcommand{\promptbox}[2]{%
\begin{tcolorbox}[enhanced jigsaw,breakable,pad at break*=1mm, title=#1,
  colback=blue!10, fontupper={\fontfamily{lmtt}\selectfont}]
  #2
\end{tcolorbox}
}

\usepackage{authblk}

\title{Improving Zero-Shot ObjectNav with Generative Communication \vspace{-0.3cm}}
\author{Vishnu Sashank Dorbala*}
\author{Vishnu Dutt Sharma*}
\author{Pratap Tokekar}
\author{Dinesh Manocha}
\affil{University of Maryland, College Park}
\affil[ ]{\small{Supplementary materials, including Code, Video, and Datasets, can be found at \url{https://gamma.umd.edu/gencom/}}}

\newcolumntype{M}[1]{>{\centering\arraybackslash}m{#1}}

\newcommand\Item[1][]{%
  \ifx\relax#1\relax  \item \else \item[#1] \fi
  \abovedisplayskip=0pt\abovedisplayshortskip=0pt~\vspace*{-\baselineskip}}

\begin{document}

\maketitle

\begin{abstract}
We propose a new method for improving zero-shot ObjectNav that aims to utilize potentially available environmental percepts for navigational assistance.
Our approach takes into account that the ground agent may have limited and sometimes obstructed view.
Our formulation encourages \textit{Generative Communication} (GC) between an assistive overhead agent with a global view containing the target object and the ground agent with an obfuscated view; both equipped with Vision-Language Models (VLMs) for vision-to-language translation. In this assisted setup, the embodied agents communicate environmental information before the ground agent executes actions towards a target. Despite the overhead agent having a global view with the target, we note a drop in performance ($-13\%$ in OSR and $-13\%$ in SPL) of a \textit{fully cooperative} assistance scheme over an unassisted baseline. In contrast, a \textit{selective} assistance scheme where the ground agent retains its independent exploratory behaviour shows a $10\%$ OSR and $7.65\%$ SPL improvement. 
To explain navigation performance, we analyze the GC for unique \textit{traits}, quantifying the presence of hallucination and cooperation. Specifically, we identify the novel linguistic trait of \textit{preemptive hallucination}  in our embodied setting, where the overhead agent assumes that the ground agent has executed an action in the dialogue when it is yet to move, and note its strong correlation with navigation performance.
We conduct real-world experiments and present some qualitative examples where we mitigate hallucinations via prompt finetuning to improve ObjectNav performance. 
\end{abstract}

\section{Introduction}
Finding and navigating to an object in an unseen environment, or \textit{ObjectNav}, is a well-studied problem in Embodied AI \cite{anderson2018evaluation, objnavsurvey1, yadav2023habitat}. To accomplish this task, an embodied agent must interpret its surroundings based on its observations and move towards areas with a higher likelihood of containing the target object. Learning-based approaches have been used for these tasks~\cite{objnavsurvey1, weihs2021visual}, but they require training and are not generalizable. Moreover, they can result in  poor out-of-domain performance, motivating development of zero-shot techniques~\cite{clipnav}. 
In zero-shot ObjectNav, an embodied ground agent navigates to a target object specified by a natural language label without any prior training or knowledge of the environment.
Several works have recently explored this variation \cite{gadre2023cows, yokoyama2023vlfm, kuang2024openfmnav, zson2, majumdar2022zson, ma2024doze, wu2024voronav}, coming up with various solutions for navigation without any fine-tuning or pre-training. This has involved using foundation models including vision-language models (VLMs) and large language models (LLMs) for commonsense reasoning and planning~\cite{sarch2023helper,padmakumar2023multimodal}. An embodied agent equipped with such models can reason about the possible locations of the target object with open-set scene understanding~\cite{dorbala2023can,shah2023navigation,pan2023langnav}. 

 \begin{figure}[ht]
    \centering
    \includegraphics[width=\linewidth, height=4.8cm]{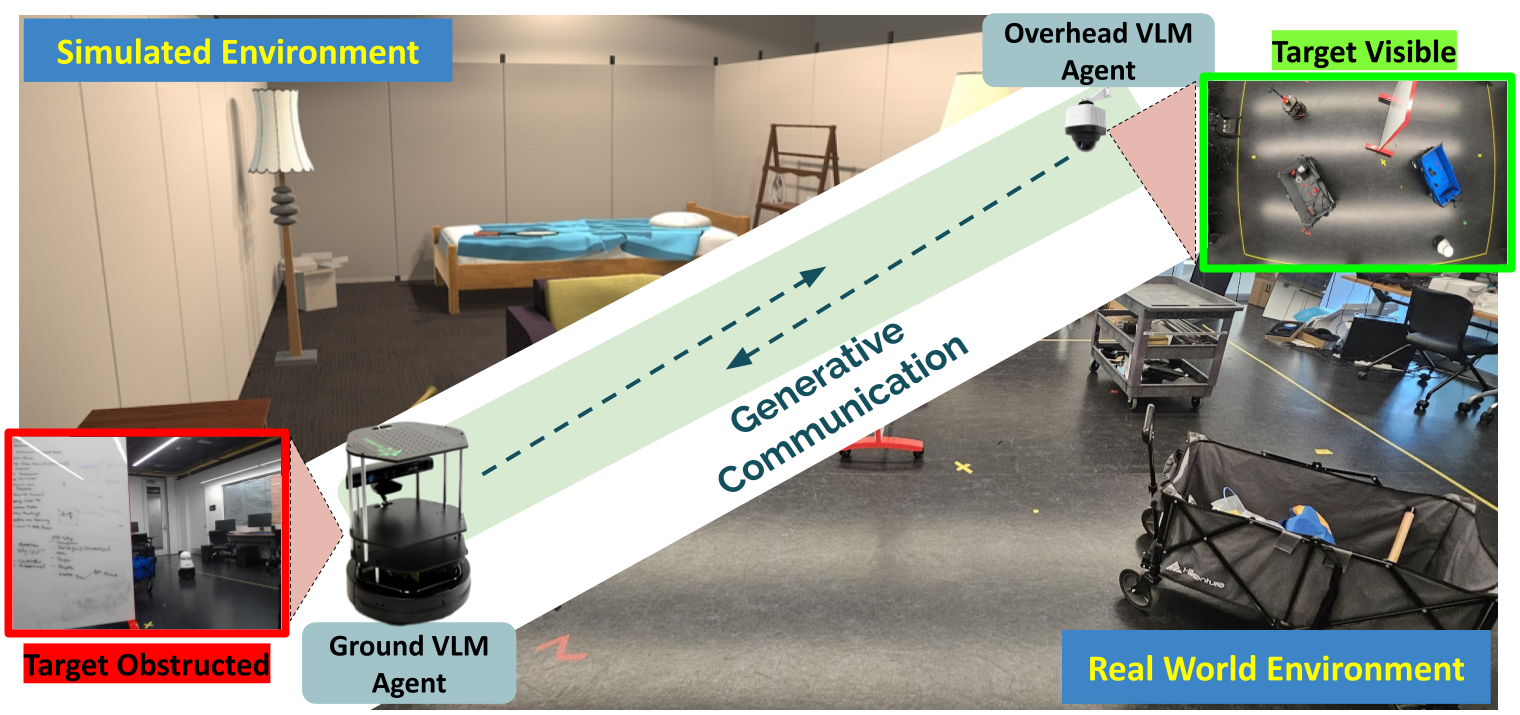}
    \caption{\textbf{Overview}: We tackle zero-shot ObjectNav in an assisted setup, where the ground agent aims to improve performance by seeking assistance from other available environmental percepts.  We consider an overhead agent (as shown) with a clear view of the target and a ground agent with an obstructed view of the target that convey environmental information to each other via freeform, unconstrained \textbf{Generative Communication (GC)}. We use GC to develop two novel \textit{assisted} navigation schemes and present results in both simulated and real-world environments, inferring that GC is useful only in a selective setup where the ground agent retains its independent exploration capability. 
    }
    \label{fig:overview}
    \vspace{-0.5cm}
\end{figure}


While several zero-shot approaches for ObjectNav have been proposed, task performance still remains poor due to the difficulty in accurately estimating a target's location in a completely unknown environment~\cite{objnavsurvey1}. This challenge is further exacerbated in single-agent setups~\cite{vlmobjnav2,vlmobjnav3} with only one agent independently exploring the environment using a local view limited by the camera field-of-view and occluding objects. To improve zero-shot performance, we consider making better use of potentially available environmental percepts, particularly considering a setup with an overhead camera having a global view containing the target, providing \textit{assistive} guidance to a ground agent via generative natural language. 

Prior work by Patel et al.~\cite{commbase1} demonstrates the benefit of an overhead view for such navigation, although with emergent communication, which requires training a communication module. Given the existence of monitoring cameras in industrial and commercial settings and the proliferation of smart home devices equipped with cameras for various applications~\cite{h2022monitoring}, it is beneficial to utilize them for assistance. This is analogous to unmanned aerial vehicles (UAVs) guiding ground robots by acting as scouts\cite{sharma2020risk, asghar2023risk, karapetyan2024ag}.

To this end, we assume a collaborative setup where a dynamic ground agent communicates with a static overhead agent seeking to improve its ObjectNav performance. 
Using the expansive view of the environment, the overhead agent is expected to guide the ground agent towards regions with high likelihood of containing the target object.
The ground agent reports the details of the environment to provide better context to the overhead agent about its own location and surroundings.
In such collaborative setups, effective communication is necessary for meaningful collaboration ~\cite{commbase1}. The translation of complex perceptual cues into a shared and sufficiently descriptive language for precise navigational actions is a necessity in this regard. We can leverage the powerful translational capabilities of VLMs~\cite{vlmsur1, vlmsur2, li2020oscar}, by equipping both ground and overhead agents with VLMs to perform vision-to-language translation.
Our goal is to develop new methods for zero-shot ObjectNav through the lens of VLM-based \textit{Generative Communication} (GC) between these agents.

A few prior works have studied collaborative setups among embodied agents, particularly dealing with emergent communication~\cite{emergentnatural, emergentsurvey1, emergentcontrol1}. These methods mainly deal with developing symbolic language with limited vocabularies for effective communication as opposed to using natural language, which offers a broader and more interpretable vocabulary. Guo et al.~\cite{guo2024embodied-mengdi} study emergent communication in embodied agents equipped with LLMs but consider a predefined leadership hierarchy on \textit{homogeneous} agents of the same type. 

\noindent {\bf Main Results:} 
We study the zero-shot ObjectNav task in an assisted setup and present two novel cooperative schemes that utilize the GC between a dynamic ground agent and a static overhead agent for generating navigation actions. Despite the overhead agent having a global view with the target in its field-of-view, we observe that a \textit{fully} cooperative approach where the ground agent follows instructions without judgement performs poorly, and that a \textit{selective} scheme which retains independent exploration of the ground agent shows improved performance. We further investigate this result by defining and analyzing linguistic \textit{traits} from the GC that have a direct impact on navigation performance. The novel components of our work include:
\begin{itemize}
    \item \textbf{Assisted Task Setup}: We tackle the zero-shot ObjectNav task in an \textit{assisted} setup, where the embodied ground agent seeks help from an overhead agent with a global view. Our results are presented in both simulator and real-world environments.

    \item \textbf{GC-Driven Navigation Schemes}: We present two novel navigation schemes (Fully and Selective Cooperation) on this setup, establishing a benchmark for zero-shot ObjectNav in an assisted setting. Our navigation schemes utilize generative communication (GC) between VLM-equipped ground and overhead agents.
    
    \item \textbf{Improving Zero-Shot ObjectNav}: On the RoboTHOR simulator \cite{robothor}, our \textit{Selective} Cooperation scheme where the ground agent exhibits self-judgment results in a performance improvement of $10\%$ in OSR and $7.65\%$ in SPL over a non-assistive baseline. In contrast, a \textit{Fully} Cooperative scheme results in a $13\%$ performance decline (both OSR and SPL), indicating that following the overhead agent's guidance as is hinders performance, \textit{despite} its global view of the target.

    \item \textbf{Novel Lingustic Traits}: We measure $4$ novel linguistic traits in the GC to help explain our navigation performance. 
    In particular, we note that the trait of \textit{Preemptive Hallucination} ($\mathcal{H_{PE}}$) is abundant in the GC, showing a strong correlation with navigation performance in the \textit{Fully} Cooperative scheme.
    

    \item \textbf{Real World Experiments}: Finally, we present qualitative real-world results using a Turtlebot as a Ground Agent and a GoPro as the Overhead Agent. We show that environment-specific fine-tuning of VLM prompts can help mitigate hallucinations for real-world success. 
\end{itemize}

\section{Related Works}
\subsection{LLMs and VLMs for Embodied Agents}
LLMs and VLMs have been used for various robotics applications.  These models are trained on internet-scale data and thus exhibit the ability to act as world models for robotic task planning and scene understanding. This has led to traditional and deep learning-based approaches being replaced by zero-shot, LLMs/VLMs-based approaches for navigation~\cite{yokoyama2023vlfm,chang2023goat,shah2023navigation,shah2022lm,xie2023reasoning}, manipulation~\cite{zhang2023bootstrap,gao2023physically,ha2023scaling} and general-purpose robots~\cite{hu2023toward}. 

LLMs can also act as a natural language to code translators, making human-robot collaboration more feasible than prior approaches. Such approaches have been explored to obtain a plan as a PDDL~\cite{silver2022pddl} or python code~\cite{liang2023code}. For VLMs, recent works have also successfully used visual prompting to help robots~\cite{nasiriany2024pivot,sathyamoorthy2024convoi} with visual cues grounded using the robot's capabilities. This has been used for effective human-robot collaboration. But it has given rise to the question: \textit{How would a team of multiple such robots behave?} Historically, the information exchanged between agents in a multi-agent team has been handcrafted~\cite{commbase1} or learned during task-specific training~\cite{zhou2022graph}. Natural language-based, generative communication for multi-agent teams of embodied agents is a fairly recent and relatively unexplored domain. Guo et al.~\cite{guo2024embodied-mengdi} evaluate this situation and show that communication leadership improves team efficiency for such LLM-powered agents. However, they focus only on a team of homogeneous agents.

\begin{figure*}[t!]
    \centering
    \includegraphics[width=\linewidth]{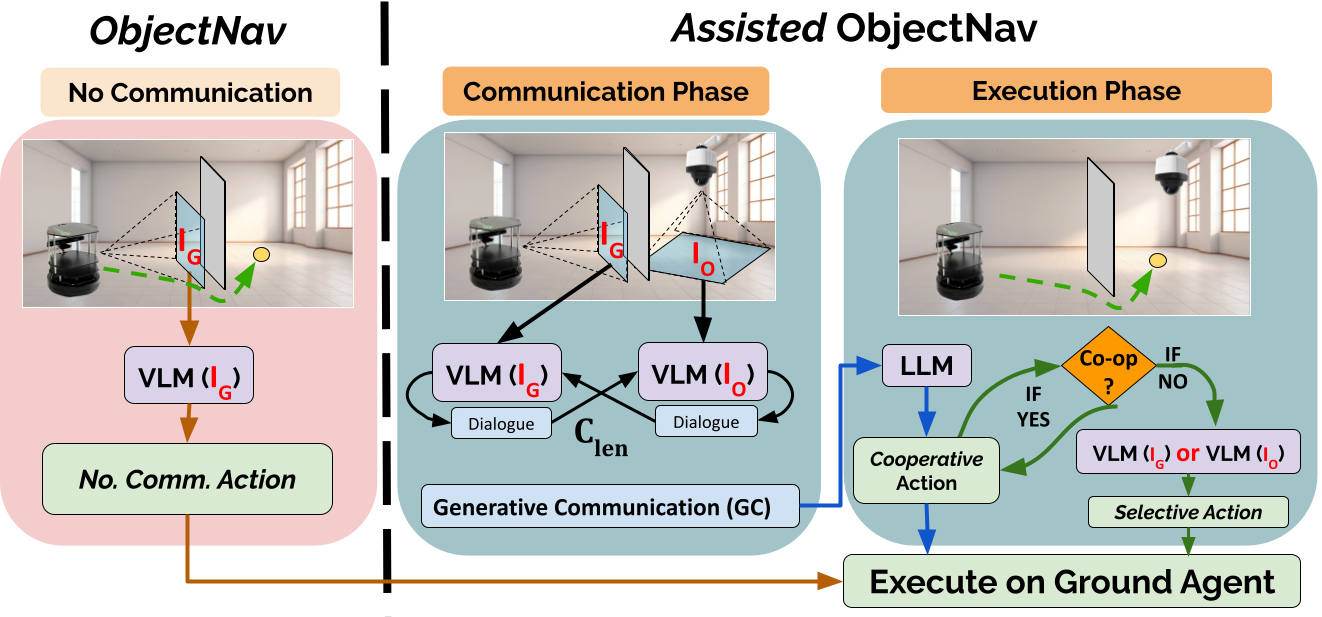}
    \caption{\textbf{Approach}: We consider 3 different setups for assisted ObjectNav on a ground agent (GA) using an overhead agent (OA). The \textbf{No Comm.} case (brown arrows) is a \textit{baseline} ObjectNav setup where the GA is prompted directly by a VLM for navigation actions for the ground agent. This is illustrated on the left. For the remaining two cases, both agents first go through a Comm. phase ($\mathcal{C}$) for a fixed number of interactions $C_{\text{len}}$. We then summarize the dialogues for decision-making. In the \textbf{Cooperative Action} case (blue arrows), we pass the Generative Communication (GC) to an LLM that predicts an action for the GA. In the \textbf{Selective Execution} case (green arrows), the GA's VLM is prompted with the suggested action and asked if it wants to cooperate with the LLM prediction. If not, it performs independent exploration like the No. Comm. case. We later analyze the dialogue generated to measure generative communication traits.}
    \label{fig:flowchart}
    \vspace{-0.5cm}
\end{figure*}

\subsection{Ground-Overhead Agent Collaboration}
A team of embodied agents with different viewpoints can enhance team capabilities by combining their respective strengths. For example, in a team of a UAV (unmanned aerial vehicle) and a UVG (unmanned ground vehicle) deployed for rescue in a disaster area, the UAV can act as a scout, observing a large area and guiding the UGV to reach affected people. The UGV can then perform long-range navigation and carry a heavy payload in the form of supplies, which a UGV may not be able to carry for a long time. Thus, effective collaboration would depend upon communication. Chen et. al. in \cite{chen2023scalable} studied natural language-based communication coming from LLMs for heterogeneous agents in various situations. Their focus, however, is on scalability, and they perform experiments in simple grid-like environments. Our work instead focuses on embodied agents in indoor environments, specifically how communication influences ObjectNav performance. Along with studying the communication length, similar to \cite{chen2023scalable}, we also develop metrics to measure hallucination in generating communication originating from LLMs and its effect on collaboration. 

\subsection{Communication in Cooperating Agents}
When multiple agents communicate with each other to accomplish a task together, they effectively engage in a referential game \cite{lazaridou2016multi}. Earlier approaches studying referential games used symbolic communication \cite{havrylov2017emergence,lazaridou2018emergence,kottur2017natural}, but with the advancements in deep learning, attention shifted to communicating with embeddings, often learned over time with reinforcement learning~\cite{lazaridou2020emergent}. The development of language models then spurred an interest in natural language-driven methods \cite{lazaridou2020multi,kim2024knowledge}. Visual signals have also been explored to enhance emergent communication \cite{evtimova2017emergent,mu2023ec2}. These works, however, involve learning to communicate, which forms the basis for emergent traits. 
Unlike 'selective communication` in most multi-agent simulation work \cite{selcom1, zhai2021decentralized} where the agent chooses which of the other agents it wishes to communicate with, in our work, we consider `selective execution', where the agent chooses whether or not to act on the information given. Would we observe such traits when an LLM or a VLM is used in a zero-shot manner? Guo et. al. in \cite{guo2024embodied-mengdi} looked at some such traits but for homogeneous agents with a predefined leadership hierarchy. We aim to identify and analyze these traits for homogeneous agents in our approach. 

Prior work by Patel et al.~\cite{commbase1} is the closest to our work, addressing the issues above by proposing a framework using limited pre-defined vocabulary acting as a black box for communication rendering limited specific to limited environments. Furthermore, they use specific types of objects (colored cylinders) as goals. In our work, we deal with the case of fully generated natural language not limited by vocabulary or environments to find objects.

\section{Approach}
\subsection{Definitions}

\noindent\textbf{Assisted ObjectNav Task}: We introduce a variation of the ObjectNav task called \textit{assisted} ObjectNav. In this task, an embodied ground agent must consider using \textit{environmental assistance} to optimize its path towards a target object. In this work, we tackle a baseline case for this task, where the ground agent receives assistance from one \textit{static} agent with a different viewpoint of the environment. In this dual-agent setup, the agents constitute a vertical hierarchy \cite{chen2023agentverse} and are equipped with VLMs for collaborative decision-making. The agents are expected to communicate and cooperate to identify and navigate towards the target. We further that the agents are always within each other's communication range and use the same type of VLM.



\noindent\textbf{Agents}: We consider two VLM-equipped agents: \\
1) A \textbf{Ground Agent (GA)} that has a \textit{local view} of the environment and can perform physical maneuvers to reach the target. This is the embodied agent physically performing the ObjectNav task. \\
2) An \textbf{Overhead Agent (OA)} that has a \textit{global view} of the environment containing the target object. This agent is stationary and aims to assist the GA reach the target.

As a baseline, we consider a no-communication setup, where the GA aims to perform ObjectNav with VLM independentally. In the assisted ObjectNav setup, our approach involves an interplay between two phases: a \textbf{Communication Phase, $\mathcal{C}$}, where both agents generate dialogue with information about their respective egocentric views, and an \textbf{Execution Phase, $\mathcal{E}$}, where the GA receives a maneuvering recommendation from the other agent and must then decide what action the ground agent must execute.
Fig. \ref{fig:overview} presents an overview of our setup for the baseline ObjectNav and the assisted ObjectNav tasks.

\subsection{No Communication Baselines}
\label{sec:no_comm}
We consider a baseline case where the GA uses its VLM to guide itself independently toward the target object. The agent processes its egocentric viewpoint along with a prompt about navigating to a target object to get a directional command to execute. The prompts used are in the Appendix.
We also present a \textbf{Random} decision-making approach where the GA randomly chooses and executes a directional command to highlight the utility of informed actions. 


\begin{figure*}[h!]
    \centering
    \begin{subfigure}[t]{0.55\textwidth}
        \centering
        \includegraphics[width=\linewidth]{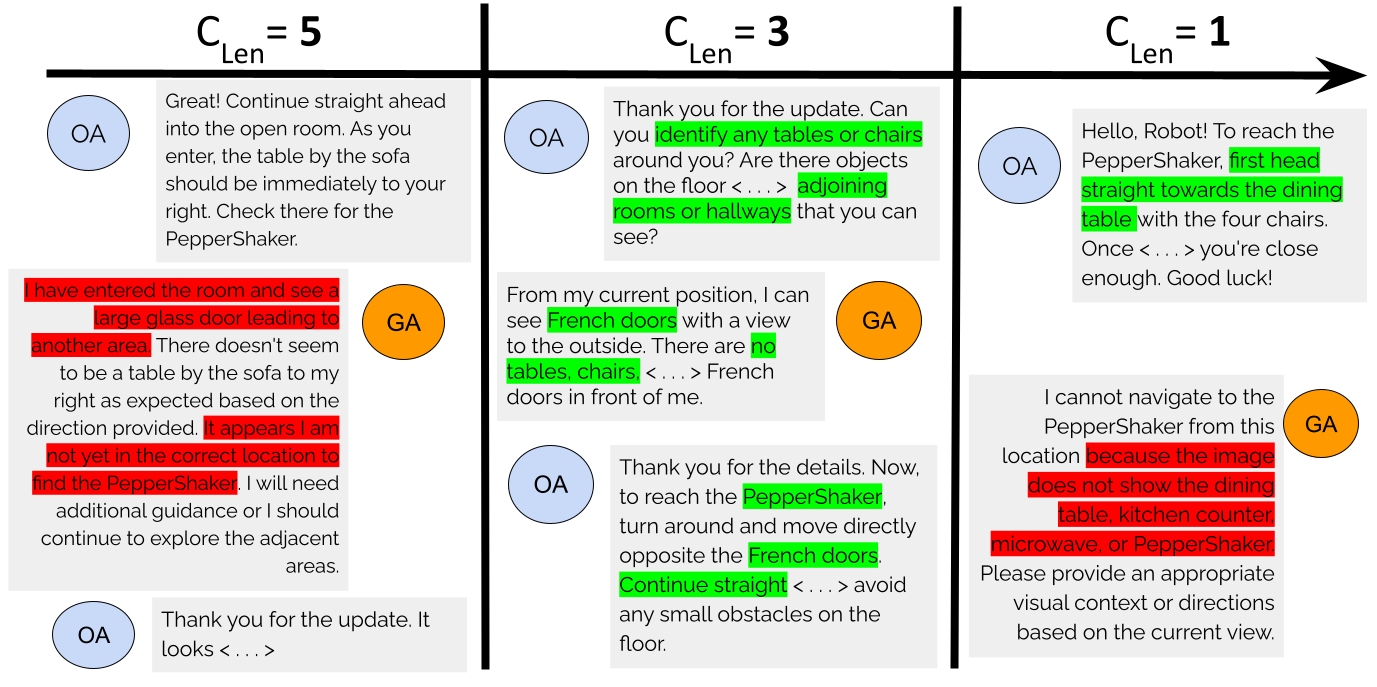}
        \caption{Dialogue Hallucinations: Different $\mathcal{C}_{Len}$ examples.}
        \label{fig:hallucination_a}
    \end{subfigure}%
    \hfill
    \begin{subfigure}[t]{0.44\textwidth}
        \centering
        \includegraphics[width=\linewidth]{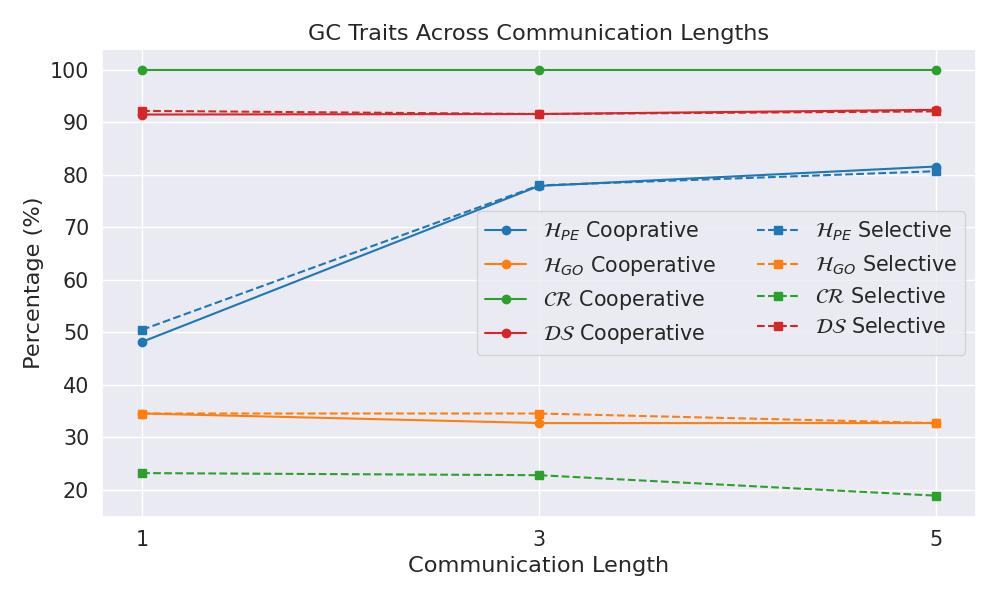}
        \caption{Comparing GC Traits across various Communication Lengths.}
        \label{fig:hallucination_b}
    \end{subfigure}
    
    \caption{\textbf{Dialogue Hallucinations} $\mathcal{H}$: We study hallucinations in dialogues to explain assisted ObjectNav performance. In the left figure, the target object is a \textit{PepperShaker}, showing examples of hallucinations at different communication lengths. The right figure compares hallucinations (\textbf{$\mathcal{H}_{PE}$}, \textbf{$\mathcal{H}_{GO}$}), Cooperation Rate $\mathcal{CR}$ and Dialogue Similarity $\mathcal{DS}$ between cooperative and selective actions. Among all the traits, notice that \textbf{$\mathcal{H}_{PE}$} stands out, getting worse as communication length increases, despite explicitly prompting for concise information.}
    \vspace{-0.5cm}
    \label{fig:combined_hallucination}
\end{figure*}

\subsection{Communication Phase ($\mathcal{C}$)}
Improving collaborative task success requires agents to both gather information about the environment with their percepts as well as effectively translate perception data to robot language. Both the GA and the OA are equipped with VLMs capable of generating natural language conversations. The GA seeks assistance from the OA and prompts the OA for a dialogue. We specify to the OA that it must communicate for a length of $C_{\text{Len}}$, asking the GA queries that would help guide it to the target object. The GA then responds to the generated dialogue, and this cycle happens back and forth for $C_{\text{Len}}$ times. At the end of the communication period, we concatenate all the dialogues into a summary containing potentially useful information about the environment for making navigation decisions. We consider this dialogue as \textbf{Generative Communication} or GC (see Fig. \ref{fig:flowchart}). 

\subsection{Execution Phase ($\mathcal{E}$)}
While successful communication protocols have traditionally been learned using RL and improve performance on the heterogeneous ObjectNav task \cite{commbase1}, to investigate the performance of GC in our setup, we consider the two zero-shot execution variants: \textbf{Cooperative Action} and \textbf{Selective Action}.

In the \textbf{Cooperative Action} setup, the agents exchange dialogue of varying lengths ($C_{\text{Len}}$) in $\mathcal{C}$. After this, in $\mathcal{E}$, the GC is passed to an LLM, which then suggests an action for the GA to take. The GA \textit{fully cooperates} with the LLM's prediction and directly executes the prescribed action.

In the \textbf{Selective Action} setup, the GA is presented with a choice on whether it wishes to cooperate with the LLM's decision. The GA is prompted with the LLM action output for a ``Yes/No'' on whether it should cooperate with the LLM's recommendation. If it decides to cooperate (Yes), then the action recommended by the LLM is executed. If the GA declines to cooperate (No), then we again prompt it to ask for its recommendation on the action to take given the GA's egocentric image. The former allows us to measure a \textit{cooperation rate} or $\mathcal{CR}$ of the agents, while the latter is similar to the No Communication phase (Sec. \ref{sec:no_comm}), where the action is taken directly from the ground VLM agent.

Both \textbf{Cooperative Action} and \textbf{Selective Action} are situations where the GA must communicate with the OA. We additionally investigate a scenario where the GA first decided whether to initiate a conversation with the OA. More details about these ablations are in the Appendix. 

\section{Generative Communication Traits}
\label{sec:GC_traits}
In this section, we look at quantifying various qualities that emerge during the VLM-based generative communication between the agents. We derive these from the dialogue generated during the communication phase and use them to assess the influence of GC on ObjectNav performance.

\subsection{Dialogue Hallucinations - $\mathcal{H}$}
Motivated by the high human-verified GPT-4 classification accuracy reported by \cite{guo2024embodied-mengdi} on dialogue analysis, we adopt a GPT-4 classifier for finding hallucinations in GC. 
We look at the following: \\
\noindent\textbf{Preemptive Actions}  ($\mathcal{H}_{PE}$): We notice that in many cases, one of the agents assumes that the GA has moved during $\mathcal{C}$, even though the GA can only move after the completion of $\mathcal{C}$ (Refer $\mathcal{C}_{Len}=5$ in Fig. \ref{fig:hallucination_a}). We refer to these as \textit{preemptive Actions}. To quantify it, we prompt a GPT-4 agent to classify whether the dialogue in an episode indicates a preemptive action hallucination. The labels obtained are averaged over all the conversations. This can be defined as: 
\[
\mathcal{H}_{PE} = \frac{1}{N}\sum^{N}_{i=1}\frac{\mathcal{L}(GC)_{i}}{k}
\]
where $\mathcal{L}(GC)$ is the GPT-based classifier for preemptive actions.
\begin{table*}[ht]
\centering
\begin{tabular}{cccrrrcc}
\toprule
 & & \multicolumn{2}{c}{\textbf{Navigation}} & \multicolumn{4}{c}{\textbf{Generative Comm. Traits}} \\
\cmidrule(lr){3-4} \cmidrule(lr){5-8}
\textbf{Comm. Length ($\mathcal{C}_{Len}$)} & \textbf{Execution} & \textbf{OSR \%} $\uparrow$ & \textbf{SPL\%} $\uparrow$  & \textbf{$\mathcal{H}_{PE}$ \%} $\downarrow$ & \textbf{$H_{GO}$ \%} $\downarrow$ &\textbf{ $CR$ \%} &\textbf{$DS$\%} \\
\midrule
\multirow{2}{*}{0} & Random (No VLM) & 15.00 & 11.69 & N/A & N/A & N/A & N/A \\
 & No Comm. \textit{Baseline} & 22.00 & 21.50 & N/A & N/A & N/A & N/A \\

\specialrule{.3em}{.05em}{.05em}
\multirow{2}{*}{1}
                    & Cooperative Action & 19.00 (\textcolor{red}{-3.0}) & 18.38 (\textcolor{red}{-3.1})& \textbf{48.20} & 34.55 & 100 & 91.5 \\
                    & \cellcolor{cyan!20} Selective Action & \cellcolor{cyan!20} 26.00 (\textcolor{teal}{+4.0}) & \cellcolor{cyan!20} 24.13 (\textcolor{teal}{+2.6}) & \cellcolor{cyan!20} 34.55 & \cellcolor{cyan!20} 42.51 & \cellcolor{cyan!20} 23.20 & \cellcolor{cyan!20} 92.2 \\ 
\midrule
\multirow{2}{*}{3}
                    & Cooperative Action & 19.00 (\textcolor{red}{-3.0}) & 17.94 (\textcolor{red}{-3.5}) & 77.90 & \textbf{32.73} & 100 & 91.6\\
                    & \cellcolor{cyan!20} Selective Action & \cellcolor{cyan!20} 24.00 (\textcolor{teal}{+2.0}) & \cellcolor{cyan!20} 22.19 (\textcolor{teal}{+0.6}) & \cellcolor{cyan!20} 78.00 & \cellcolor{cyan!20} 34.55 & \cellcolor{cyan!20} 22.80 & \cellcolor{cyan!20} 91.6 \\
\midrule
\multirow{2}{*}{5}
                    & Cooperative Action & 9.00 (\textcolor{red}{-13.0}) & 8.50 (\textcolor{red}{-13.0}) & 81.60 & \textbf{32.73} & 100 & 92.4 \\
                    & \cellcolor{cyan!20} Selective Action & \cellcolor{cyan!20} \textbf{32.00} (\textcolor{teal}{\textbf{+10.0}}) & \cellcolor{cyan!20} \textbf{29.15} (\textcolor{teal}{\textbf{+8.0}}) & \cellcolor{cyan!20} 80.70 & \cellcolor{cyan!20} \textbf{32.73}  & \cellcolor{cyan!20} 18.90 & \cellcolor{cyan!20} 92.1 \\
\bottomrule
\end{tabular}
\caption{\textbf{ObjectNav Results and GC Traits}: We measure OSR and SPL across different $\mathcal{C}_{Len}$. Note the improved Navigation performance both on SPL and OSR in the Execution cases with Selective Actions. The GA in these cases retains its \textit{independent} navigation capability, and is given a choice on whether to follow the recommended action or not. On the GC traits, we infer that VLM hallucination gets progressively worse in assuming preemptive agent motion ($\mathcal{H}_{PE}$) with increasing $\mathcal{C}_{Len}$, which could be attributed to cascading \cite{cascadingcode}. The decreasing cooperation rate $\mathcal{CR}$ and increasing $\mathcal{H}_{PE}$ Selective Action cases further suggest that when given a choice to cooperate, the GA would not trust actions produced from hallucinated summaries, explaining the superior performance.}
\label{tab:sim_objectnav}
\vspace{-0.5cm}
\end{table*}


We also look for perceptual hallucinations of objects not present in the environment (or \textit{ghost objects}) $\mathcal{H}_{GO}$ in the dialogue. This is quantified as the overlap between a vocabulary consisting of objects in the environment and the dialogue generated. More details on $\mathcal{H}_{GO}$ and the prompts we use for classification are in the appendix.

\subsection{Communication Length - $\mathcal{C}_{Len}$}
The length of communication, $\mathcal{C}_{Len}$, sets the limit on the amount of information that can be exchanged between the two agents (especially with a fixed number of generated tokens). We ablate on $C_{Len}$ of $1$, $3$, and $5$ for our approach and study its effect on other metrics.

\subsection{Cooperation Rate - $CR$}
In the \textit{Selective Action} setting, the GA may choose to not comply with the LLM recommendation, instead choosing to act independently. We measure this cooperation rate as the number of times the agent chooses to cooperate, averaged over an episode. 
\[
\mathcal{CR} = \frac{1}{N}\sum^{N}_{i=1}\frac{c_{i}}{k},
\]
where $c_{i}$ is the number of times the GA decided to cooperate in the $i^{\text{th}}$ episode with $k$ steps, and $N$ is the total episode count.

A lower cooperation rate would indicate more independent exploration, with the GA choosing to ignore the recommended actions. Further, this could be an indirect indicator for hallucinated communication resulting in unfeasible actions.




In addition to these, we also quantify the dialogue similarity, $\mathcal{DS}$, as the cosine similarity between the BERT embeddings of dialogues across episodes at each step. Details about this can be found in the appendix.

\section{Experiments \& Results}
We measure ObjectNav and GC performance via experiments in RoboTHOR as well as a real-world setup. RoboTHOR allows us to conduct large-scale experiments in a simulated environment, allowing us to measure task performance objectively via Oracle Success Rate \textbf{(OSR)} and Success Weighted by Path Length \textbf{(SPL)} as defined by \cite{anderson2018evaluation}. In the real world, we present some unique qualitative findings on differences in the GC and visual localization of the overhead agent, in contrast with simulator experiments. In both cases, we compute and validate the previously defined GC Traits.

\subsection{Simulation: RoboThor}
\label{subsec:simualtion_setup}
\noindent\textbf{Setup}:
We use 100 rooms from the training split of ProcTHOR \cite{deitke2022️} for our experiments. In each room, we place a LoCoBot robot with an RGB camera as the GA and add an overhead camera. The overhead camera, our OA, is located at a height of 7.25m. We crop the images from the overhead camera to minimize the views outside the house before passing them to the VLM. We use \textit{GPT-4-turbo} as the VLM in all the experiments. 

At the start of each episode, the GA spawns at the default location for the room in the ProcTHOR setup. As the target object, we randomly select an object from the scene such that only one instance of it is available. We take 10 action steps during each episode and keep track of its outcome. \DUTT{We use the standard discrete action primitives used in the embodied ObjectNav tasks~\cite{weihs2021visual,anderson2018evaluation}.} Hence, the ground robot is allowed to take one of the following actions: \textit{MoveAhead}, \textit{MoveBack}, \textit{RotateLeft}, \textit{RotateRight}, or \textit{DoNothing}. All the movements result in the GA moving by 0.25m in the indicated direction. The rotation actions result in $90^\circ$ rotation while staying at the same place. \textit{DoNothing} lets the GA stay where it is without any change.

During the communication phase, $\mathcal{C}$, the agents use their respective camera views and talk to each other to devise a plan to reach the target object. Then we share generated communication between the agents, the past action, and the result of it (success, collision, etc.) with the decision-making agent during the execution phase $\mathcal{E}$. In a non-cooperative setup, the agent uses its camera view to decide the appropriate action to reach the target object.

\noindent\textbf{Navigation Inferences}:
Table \ref{tab:sim_objectnav} presents simulation results on Zero-Shot ObjectNav and the GC traits obtained. We make two inferences: \textbf{1)} An agent using Selective Actions performs far better ($10\%$ OSR over baseline) than one that follows OA suggestions without judgement (Cooperative Action). This setup performs even worse than if the agent were not communicating at all. This holds true across all $\mathcal{C}_{Len}$, and is especially worse on $\mathcal{C}_{Len}$ 5 with a dip of $13\%$. This indicates that even when assisted, an agent must be provided with the choice to reject suggestions and independently explore to improve performance. \textbf{2)} There is a drastic drop in Cooperative Action performance at $\mathcal{C}_{Len} = 5$, especially with SPL. We infer that actions predicted from longer communications tend to be worse and negatively impact the performance of a fully cooperative agent.

\begin{figure*}[t!]
    \centering
    \includegraphics[width=0.90\linewidth]{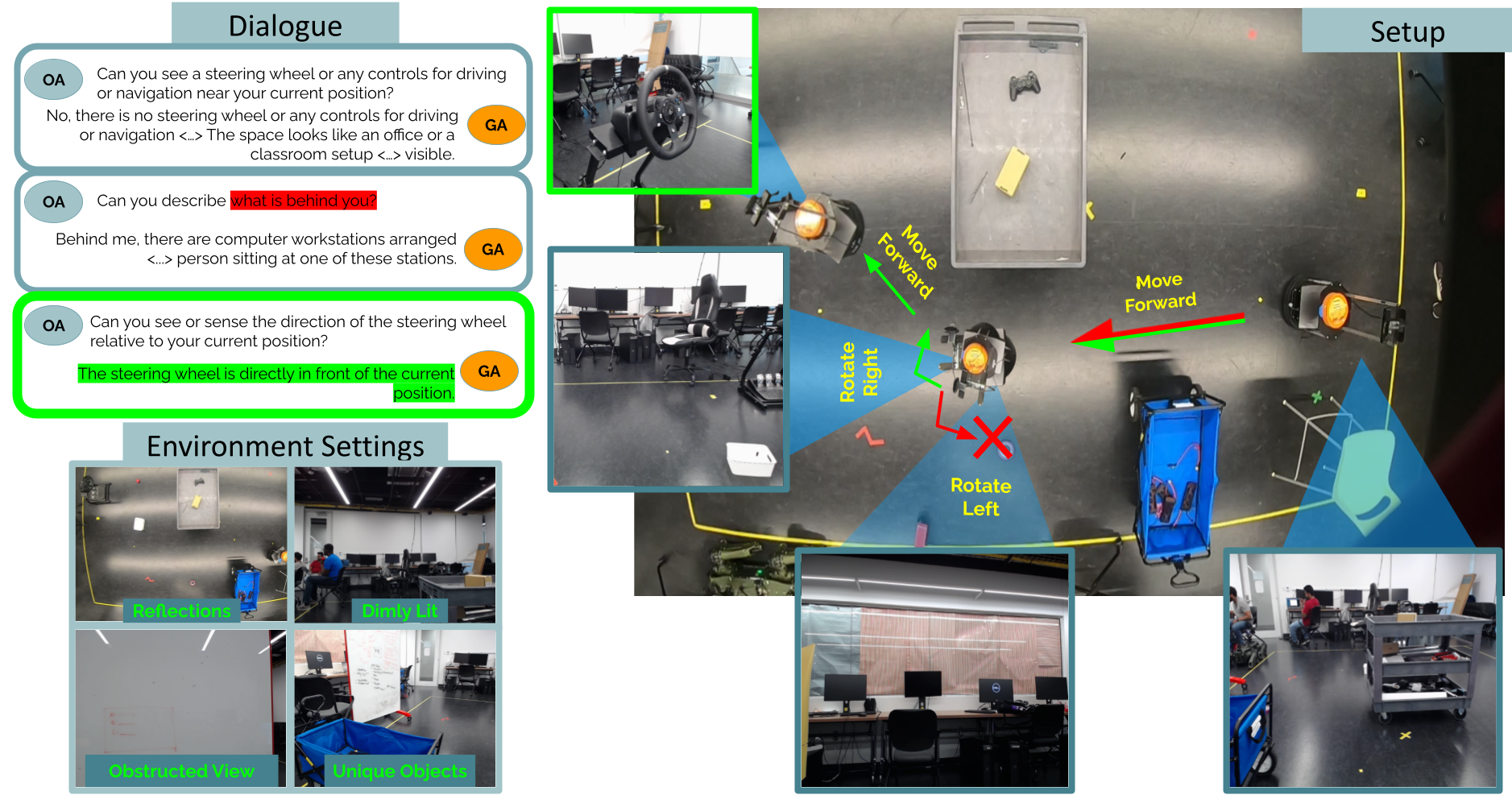}
    \caption{\textbf{Real World Experiments}: We carry out a real-world experiment with a Turtlebot as a Ground Agent (GA) and a GoPro camera mounted to the roof as an Overhead Agent (OA) in various environment settings. Note the incorrect action taken in the cooperative execution case (red arrows) in comparison to the selective case (green arrows). The actions predicted are in yellow. Section \ref{sec:quant_rw} below discusses various hallucinations with different environment settings we encounter and how we finetune VLM prompts for better results.}
    \label{fig:rw_setup}
    \vspace{-0.5cm}
\end{figure*}

\noindent\textbf{Hallucination Inferences}: Fig. \ref{fig:hallucination_b} shows the trends GC traits across comm. lengths. Note the increasing preemptive hallucinations $\mathcal{H}_{PE}$ with increasing $\mathcal{C}_{Len}$. We also find that $\mathcal{H}_{GO}$ remains almost constant across $\mathcal{C}_{Len}$.
This indicates that while GPT-4V is consistent with regards to describing the content in images, it tends to hallucinate a lot more when it comes to generating \textit{preemptive actions}. Examples of such conversations are provided in Fig. \ref{fig:hallucination_a}. Despite being explicitly prompted to converse only about visual elements during the conversation and being told that the agent cannot execute actions, GPT-4 tends to hallucinate in assuming that the agent has moved.

Further, note the decreasing Cooperation Rate, $\mathcal{CR}$, with increasing hallucinations. This relation explains the improved ObjectNav performance, showing that an agent with Selective Actions cooperates less with hallucinated instruction. We also found that the dialogue similarity $\mathcal{DS}$ is almost similar across all approaches and as such is not a good metric to evaluate the dialogue sensitivity. Our preemptive Hallucination metric $\mathcal{H}_{PE}$ is far more informative in allowing us to gauge the performances of these approaches. 

\subsection{Qualitative Real World Experiments}
\label{sec:quant_rw}
\noindent\textbf{Setup}: We recreate our dual agent setup in the real world using Turtlebot 2 as the ground agent and a GoPro Hero 7 as the overhead agent (Fig. \ref{fig:rw_setup}). We use GPT-4V as our VLM, noting the very poor performance of other VLMs, particularly InstructBLIP and LLaVA. More details on these results are in the appendix. 
We arrange objects in our lab environment as obstacles, asking the agent to find unique objects such as \textit{white basket}, \textit{whiteboard}, \textit{yellow chair}, etc., with different lighting settings, humans and clutter.\\
\noindent\textbf{Poor VLM Localization}: We notice that GPT-4 often misidentifies the position of the robot or target from an overhead view \cite{topviewvlm}, resulting in incorrect guidance to the GA.
Environmental artifacts may also confound the VLM, hindering performance. One such instance we encountered was where the OA thought the reflection of light on the floor was \textit{white tables}, leading to incorrect suggestions to the GA. Also, while looking for a \textit{white basket} as the target, the VLM sometimes misinterpreted the reflections as the target.
These are illustrated in Fig. \ref{fig:vlm-loc1} and Fig. \ref{fig:vlm-loc2}, with more examples in the appendix.

\begin{figure}[h!]
    \centering
    \includegraphics[width=\linewidth]{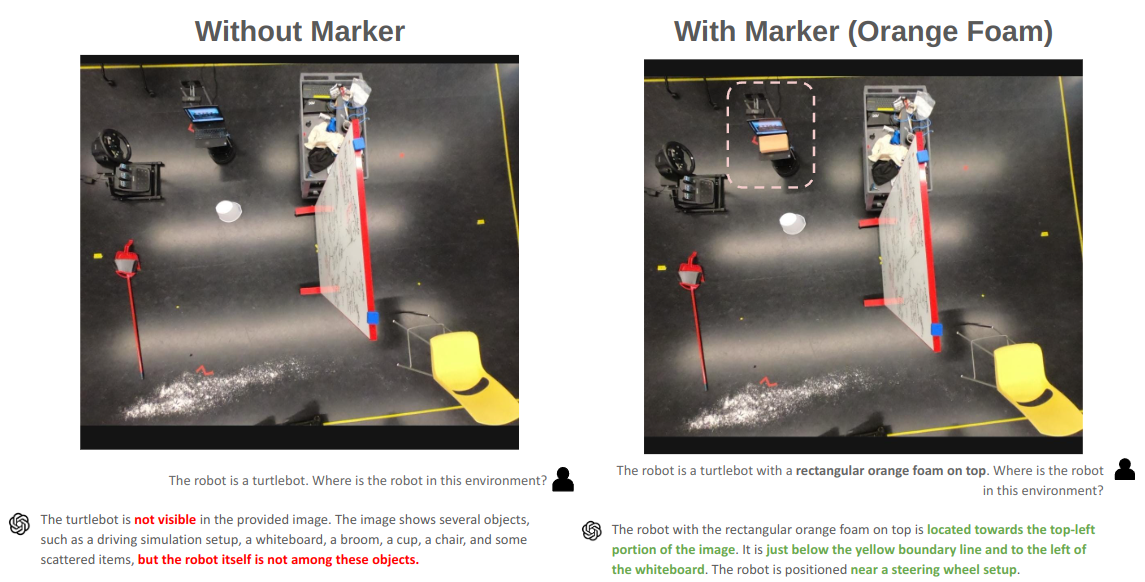}
    \caption{\textbf{Top-View Localization}: Notice the poor localization of GPT-4V on an overhead image when asked to identify the location of the robot. Adding an orange marker on the robot helps alleviate this issue.}
    \label{fig:vlm-loc1}
\end{figure}

\begin{figure}[h!]
    \centering
    \includegraphics[width=\linewidth]{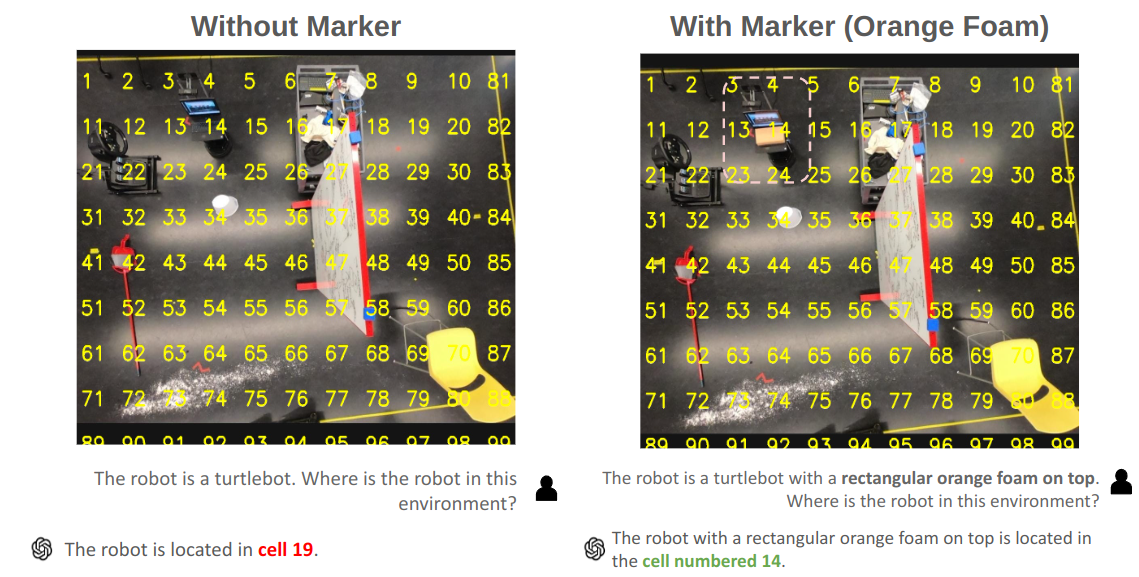}
    \caption{\textbf{Interpreting VLM Localization}: We add numbers to the image taking inspiration from recent works \cite{sathyamoorthy2024convoi, nasiriany2024pivot} to better interpret VLM localization. On the left image, note VLM prediction error in predicting the wrong cell, which is mitigated by adding an orange marker on the turtlebot.}
    \label{fig:vlm-loc2}
\end{figure}

We seek to mitigate such hallucinations via prompt finetuning with prior knowledge of the environment.
\begin{itemize}
    \item \textbf{Prompt Finetuning}: We conduct experiments in various environment settings, including dimly lit rooms with humans, clutter, and unique objects. For improving VLM localization, we put an orange box on top of the robot, as shown in Fig.~\ref{fig:rw_setup}, and modify the OA prompt accordingly. Similarly, we describe the presence of light reflections in the prompts, leading to better decision-making. We share more detailed discussion on these experiments in the appendix.

    \item \textbf{Selective vs Fully Cooperative}: Similar to the simulation experiments, we find that selective actions result in better navigation compared to fully cooperative ones. In Fig. \ref{fig:rw_setup} note the faulty Cooperative Action in red, suggesting the GA to rotate left while looking for a \textit{steering wheel} in the environment. In the Selective Action case, notice the GA ignores this suggestion. This could be due to a part of \textit{steering wheel} being visible on the right side of the image, prompting it to ignore the suggestion, rotate right, and finally reach the target.
    
    \end{itemize}

\section{Conclusion}
We present an approach to improve zero-shot ObjectNav using assistance from potentially available environmental percepts. Using VLM-equipped ground and overhead agents, we develop two generative communication (GC) driven schemes for cooperative navigation. Despite the overhead agent having a global view with the target in it, we observe a dip in performance when generated instructions and actions are followed without judgment. In contrast, our selective cooperation approach that retains the agent's independent exploration capability while selectively choosing guidance improves zero-shot ObjectNav performance in OSR and SPL; indicating the importance of an agent retaining its exploration capability while seeking external help. We explain navigation results by defining GC traits, noting the strong negative correlation of `\textit{preemptive hallucinations}' during communication, and present qualitative results for mitigating such hallucinations with environment-specific prompt fine-tuning to enhance zero-shot ObjectNav in the real-world. We hope our established benchmarks on assistive zero-shot ObjectNav will lead towards the development of more robust methods for cooperative embodied navigation. Our approach is limited to a one-on-one dialogue between two agents. To make better use of environmental percepts, future work will consider generalizing our scheme to multiple overhead and ground agents. This would require arbitration between multiple VLM-based agents to reach a consensus on what information is potentially useful for improving zero-shot performance.




{\small
\bibliographystyle{IEEEtran}
\bibliography{refs}

\begin{thebibliography}{10}
\providecommand{\url}[1]{#1}
\csname url@samestyle\endcsname
\providecommand{\newblock}{\relax}
\providecommand{\bibinfo}[2]{#2}
\providecommand{\BIBentrySTDinterwordspacing}{\spaceskip=0pt\relax}
\providecommand{\BIBentryALTinterwordstretchfactor}{4}
\providecommand{\BIBentryALTinterwordspacing}{\spaceskip=\fontdimen2\font plus
\BIBentryALTinterwordstretchfactor\fontdimen3\font minus \fontdimen4\font\relax}
\providecommand{\BIBforeignlanguage}[2]{{%
\expandafter\ifx\csname l@#1\endcsname\relax
\typeout{** WARNING: IEEEtran.bst: No hyphenation pattern has been}%
\typeout{** loaded for the language `#1'. Using the pattern for}%
\typeout{** the default language instead.}%
\else
\language=\csname l@#1\endcsname
\fi
#2}}
\providecommand{\BIBdecl}{\relax}
\BIBdecl

\bibitem{anderson2018evaluation}
P.~Anderson, A.~Chang, D.~S. Chaplot, A.~Dosovitskiy, S.~Gupta, V.~Koltun, J.~Kosecka, J.~Malik, R.~Mottaghi, M.~Savva \emph{et~al.}, ``On evaluation of embodied navigation agents,'' \emph{arXiv preprint arXiv:1807.06757}, 2018.

\bibitem{objnavsurvey1}
J.~Sun, J.~Wu, Z.~Ji, and Y.-K. Lai, ``A survey of object goal navigation,'' \emph{IEEE Transactions on Automation Science and Engineering}, 2024.

\bibitem{yadav2023habitat}
K.~Yadav, R.~Ramrakhya, S.~K. Ramakrishnan, T.~Gervet, J.~Turner, A.~Gokaslan, N.~Maestre, A.~X. Chang, D.~Batra, M.~Savva \emph{et~al.}, ``Habitat-matterport 3d semantics dataset,'' in \emph{Proceedings of the IEEE/CVF Conference on Computer Vision and Pattern Recognition}, 2023, pp. 4927--4936.

\bibitem{weihs2021visual}
L.~Weihs, M.~Deitke, A.~Kembhavi, and R.~Mottaghi, ``Visual room rearrangement,'' in \emph{Proceedings of the IEEE/CVF conference on computer vision and pattern recognition}, 2021, pp. 5922--5931.

\bibitem{clipnav}
V.~S. Dorbala, G.~Sigurdsson, R.~Piramuthu, J.~Thomason, and G.~S. Sukhatme, ``Clip-nav: Using clip for zero-shot vision-and-language navigation,'' \emph{arXiv preprint arXiv:2211.16649}, 2022.

\bibitem{gadre2023cows}
S.~Y. Gadre, M.~Wortsman, G.~Ilharco, L.~Schmidt, and S.~Song, ``Cows on pasture: Baselines and benchmarks for language-driven zero-shot object navigation,'' in \emph{Proceedings of the IEEE/CVF Conference on Computer Vision and Pattern Recognition}, 2023, pp. 23\,171--23\,181.

\bibitem{yokoyama2023vlfm}
N.~H. Yokoyama, S.~Ha, D.~Batra, J.~Wang, and B.~Bucher, ``Vlfm: Vision-language frontier maps for zero-shot semantic navigation,'' in \emph{2nd Workshop on Language and Robot Learning: Language as Grounding}, 2023.

\bibitem{kuang2024openfmnav}
Y.~Kuang, H.~Lin, and M.~Jiang, ``Openfmnav: Towards open-set zero-shot object navigation via vision-language foundation models,'' \emph{arXiv preprint arXiv:2402.10670}, 2024.

\bibitem{zson2}
W.~Cai, S.~Huang, G.~Cheng, Y.~Long, P.~Gao, C.~Sun, and H.~Dong, ``Bridging zero-shot object navigation and foundation models through pixel-guided navigation skill,'' in \emph{2024 IEEE International Conference on Robotics and Automation (ICRA)}.\hskip 1em plus 0.5em minus 0.4em\relax IEEE, 2024, pp. 5228--5234.

\bibitem{majumdar2022zson}
A.~Majumdar, G.~Aggarwal, B.~Devnani, J.~Hoffman, and D.~Batra, ``Zson: Zero-shot object-goal navigation using multimodal goal embeddings,'' \emph{Advances in Neural Information Processing Systems}, vol.~35, pp. 32\,340--32\,352, 2022.

\bibitem{ma2024doze}
J.~Ma, H.~Dai, Y.~Mu, P.~Wu, H.~Wang, X.~Chi, Y.~Fei, S.~Zhang, and C.~Liu, ``Doze: A dataset for open-vocabulary zero-shot object navigation in dynamic environments,'' \emph{arXiv preprint arXiv:2402.19007}, 2024.

\bibitem{wu2024voronav}
P.~Wu, Y.~Mu, B.~Wu, Y.~Hou, J.~Ma, S.~Zhang, and C.~Liu, ``Voronav: Voronoi-based zero-shot object navigation with large language model,'' \emph{arXiv preprint arXiv:2401.02695}, 2024.

\bibitem{sarch2023helper}
G.~Sarch, Y.~Wu, M.~Tarr, and K.~Fragkiadaki, ``Open-ended instructable embodied agents with memory-augmented large language models,'' in \emph{Findings of the Association for Computational Linguistics: EMNLP 2023}, 2023.

\bibitem{padmakumar2023multimodal}
A.~Padmakumar, M.~Inan, S.~Gella, P.~L. Lange, and D.~Hakkani-Tur, ``Multimodal embodied plan prediction augmented with synthetic embodied dialogue,'' in \emph{Proceedings of the 2023 Conference on Empirical Methods in Natural Language Processing}, 2023, pp. 6114--6131.

\bibitem{dorbala2023can}
V.~S. Dorbala, J.~F. Mullen~Jr, and D.~Manocha, ``Can an embodied agent find your “cat-shaped mug”? llm-based zero-shot object navigation,'' \emph{IEEE Robotics and Automation Letters}, 2023.

\bibitem{shah2023navigation}
D.~Shah, M.~R. Equi, B.~Osi{\'n}ski, F.~Xia, B.~Ichter, and S.~Levine, ``Navigation with large language models: Semantic guesswork as a heuristic for planning,'' in \emph{Conference on Robot Learning}.\hskip 1em plus 0.5em minus 0.4em\relax PMLR, 2023, pp. 2683--2699.

\bibitem{pan2023langnav}
B.~Pan, R.~Panda, S.~Jin, R.~Feris, A.~Oliva, P.~Isola, and Y.~Kim, ``Langnav: Language as a perceptual representation for navigation,'' \emph{arXiv preprint arXiv:2310.07889}, 2023.

\bibitem{vlmobjnav2}
L.~Zhang, Q.~Zhang, H.~Wang, E.~Xiao, Z.~Jiang, H.~Chen, and R.~Xu, ``Trihelper: Zero-shot object navigation with dynamic assistance,'' \emph{arXiv preprint arXiv:2403.15223}, 2024.

\bibitem{vlmobjnav3}
T.~Guan, Y.~Yang, H.~Cheng, M.~Lin, R.~Kim, R.~Madhivanan, A.~Sen, and D.~Manocha, ``Loc-zson: Language-driven object-centric zero-shot object retrieval and navigation,'' \emph{arXiv preprint arXiv:2405.05363}, 2024.

\bibitem{commbase1}
S.~Patel, S.~Wani, U.~Jain, A.~G. Schwing, S.~Lazebnik, M.~Savva, and A.~X. Chang, ``Interpretation of emergent communication in heterogeneous collaborative embodied agents,'' in \emph{Proceedings of the IEEE/CVF International Conference on Computer Vision}, 2021, pp. 15\,953--15\,963.

\bibitem{h2022monitoring}
N.~H.~Tan, R.~Y.~Wong, A.~Desjardins, S.~A.~Munson, and J.~Pierce, ``Monitoring pets, deterring intruders, and casually spying on neighbors: Everyday uses of smart home cameras,'' in \emph{Proceedings of the 2022 CHI Conference on Human Factors in Computing Systems}, 2022, pp. 1--25.

\bibitem{sharma2020risk}
V.~Sharma, M.~Toubeh, L.~Zhou, and P.~Tokekar, ``Risk-aware planning and assignment for ground vehicles using uncertain perception from aerial vehicles,'' in \emph{Proceedings of the IEEE/RSJ International Conference on Intelligent Robots and Systems (IROS)}, 2020.

\bibitem{asghar2023risk}
A.~B. Asghar, G.~Shi, N.~Karapetyan, J.~Humann, J.-P. Reddinger, J.~Dotterweich, and P.~Tokekar, ``Risk-aware recharging rendezvous for a collaborative team of uavs and ugvs,'' in \emph{2023 IEEE International Conference on Robotics and Automation (ICRA)}.\hskip 1em plus 0.5em minus 0.4em\relax IEEE, 2023, pp. 5544--5550.

\bibitem{karapetyan2024ag}
N.~Karapetyan, A.~B. Asghar, A.~Bhaskar, G.~Shi, D.~Manocha, and P.~Tokekar, ``Ag-cvg: Coverage planning with a mobile recharging ugv and an energy-constrained uav,'' in \emph{2024 IEEE International Conference on Robotics and Automation (ICRA)}.\hskip 1em plus 0.5em minus 0.4em\relax IEEE, 2024, pp. 2617--2623.

\bibitem{vlmsur1}
F.-L. Chen, D.-Z. Zhang, M.-L. Han, X.-Y. Chen, J.~Shi, S.~Xu, and B.~Xu, ``Vlp: A survey on vision-language pre-training,'' \emph{Machine Intelligence Research}, vol.~20, no.~1, pp. 38--56, 2023.

\bibitem{vlmsur2}
Y.~Du, Z.~Liu, J.~Li, and W.~X. Zhao, ``A survey of vision-language pre-trained models,'' \emph{arXiv preprint arXiv:2202.10936}, 2022.

\bibitem{li2020oscar}
X.~Li, X.~Yin, C.~Li, P.~Zhang, X.~Hu, L.~Zhang, L.~Wang, H.~Hu, L.~Dong, F.~Wei \emph{et~al.}, ``Oscar: Object-semantics aligned pre-training for vision-language tasks,'' in \emph{Computer Vision--ECCV 2020: 16th European Conference, Glasgow, UK, August 23--28, 2020, Proceedings, Part XXX 16}.\hskip 1em plus 0.5em minus 0.4em\relax Springer, 2020, pp. 121--137.

\bibitem{emergentnatural}
S.~Kottur, J.~M. Moura, S.~Lee, and D.~Batra, ``Natural language does not emerge'naturally'in multi-agent dialog,'' \emph{arXiv preprint arXiv:1706.08502}, 2017.

\bibitem{emergentsurvey1}
A.~Lazaridou and M.~Baroni, ``Emergent multi-agent communication in the deep learning era,'' \emph{arXiv preprint arXiv:2006.02419}, 2020.

\bibitem{emergentcontrol1}
Y.~Mu, S.~Yao, M.~Ding, P.~Luo, and C.~Gan, ``Ec2: Emergent communication for embodied control,'' in \emph{Proceedings of the IEEE/CVF Conference on Computer Vision and Pattern Recognition}, 2023, pp. 6704--6714.

\bibitem{guo2024embodied-mengdi}
X.~Guo, K.~Huang, J.~Liu, W.~Fan, N.~V{\'e}lez, Q.~Wu, H.~Wang, T.~L. Griffiths, and M.~Wang, ``Embodied llm agents learn to cooperate in organized teams,'' \emph{arXiv preprint arXiv:2403.12482}, 2024.

\bibitem{robothor}
M.~Deitke, W.~Han, A.~Herrasti, A.~Kembhavi, E.~Kolve, R.~Mottaghi, J.~Salvador, D.~Schwenk, E.~VanderBilt, M.~Wallingford \emph{et~al.}, ``Robothor: An open simulation-to-real embodied ai platform,'' in \emph{Proceedings of the IEEE/CVF conference on computer vision and pattern recognition}, 2020, pp. 3164--3174.

\bibitem{chang2023goat}
M.~Chang, T.~Gervet, M.~Khanna, S.~Yenamandra, D.~Shah, S.~Y. Min, K.~Shah, C.~Paxton, S.~Gupta, D.~Batra \emph{et~al.}, ``Goat: Go to any thing,'' \emph{arXiv preprint arXiv:2311.06430}, 2023.

\bibitem{shah2022lm}
D.~Shah, B.~Osinski, B.~Ichter, and S.~Levine, ``Lm-nav: Robotic navigation with large pre-trained models of language,'' \emph{Vision, and Action}, 2022.

\bibitem{xie2023reasoning}
Q.~Xie, T.~Zhang, K.~Xu, M.~Johnson-Roberson, and Y.~Bisk, ``Reasoning about the unseen for efficient outdoor object navigation,'' \emph{arXiv preprint arXiv:2309.10103}, 2023.

\bibitem{zhang2023bootstrap}
J.~Zhang, J.~Zhang, K.~Pertsch, Z.~Liu, X.~Ren, M.~Chang, S.-H. Sun, and J.~J. Lim, ``Bootstrap your own skills: Learning to solve new tasks with large language model guidance,'' \emph{arXiv preprint arXiv:2310.10021}, 2023.

\bibitem{gao2023physically}
J.~Gao, B.~Sarkar, F.~Xia, T.~Xiao, J.~Wu, B.~Ichter, A.~Majumdar, and D.~Sadigh, ``Physically grounded vision-language models for robotic manipulation,'' \emph{arXiv preprint arXiv:2309.02561}, 2023.

\bibitem{ha2023scaling}
H.~Ha, P.~Florence, and S.~Song, ``Scaling up and distilling down: Language-guided robot skill acquisition,'' in \emph{Conference on Robot Learning}.\hskip 1em plus 0.5em minus 0.4em\relax PMLR, 2023, pp. 3766--3777.

\bibitem{hu2023toward}
Y.~Hu, Q.~Xie, V.~Jain, J.~Francis, J.~Patrikar, N.~Keetha, S.~Kim, Y.~Xie, T.~Zhang, Z.~Zhao \emph{et~al.}, ``Toward general-purpose robots via foundation models: A survey and meta-analysis,'' \emph{arXiv preprint arXiv:2312.08782}, 2023.

\bibitem{silver2022pddl}
T.~Silver, V.~Hariprasad, R.~S. Shuttleworth, N.~Kumar, T.~Lozano-P{\'e}rez, and L.~P. Kaelbling, ``Pddl planning with pretrained large language models,'' in \emph{NeurIPS 2022 foundation models for decision making workshop}, 2022.

\bibitem{liang2023code}
J.~Liang, W.~Huang, F.~Xia, P.~Xu, K.~Hausman, B.~Ichter, P.~Florence, and A.~Zeng, ``Code as policies: Language model programs for embodied control,'' in \emph{2023 IEEE International Conference on Robotics and Automation (ICRA)}.\hskip 1em plus 0.5em minus 0.4em\relax IEEE, 2023, pp. 9493--9500.

\bibitem{nasiriany2024pivot}
S.~Nasiriany, F.~Xia, W.~Yu, T.~Xiao, J.~Liang, I.~Dasgupta, A.~Xie, D.~Driess, A.~Wahid, Z.~Xu \emph{et~al.}, ``Pivot: Iterative visual prompting elicits actionable knowledge for vlms,'' \emph{arXiv preprint arXiv:2402.07872}, 2024.

\bibitem{sathyamoorthy2024convoi}
A.~J. Sathyamoorthy, K.~Weerakoon, M.~Elnoor, A.~Zore, B.~Ichter, F.~Xia, J.~Tan, W.~Yu, and D.~Manocha, ``Convoi: Context-aware navigation using vision language models in outdoor and indoor environments,'' \emph{arXiv preprint arXiv:2403.15637}, 2024.

\bibitem{zhou2022graph}
L.~Zhou, V.~D. Sharma, Q.~Li, A.~Prorok, A.~Ribeiro, P.~Tokekar, and V.~Kumar, ``Graph neural networks for decentralized multi-robot target tracking,'' in \emph{2022 IEEE International Symposium on Safety, Security, and Rescue Robotics (SSRR)}.\hskip 1em plus 0.5em minus 0.4em\relax IEEE, 2022, pp. 195--202.

\bibitem{chen2023scalable}
Y.~Chen, J.~Arkin, Y.~Zhang, N.~Roy, and C.~Fan, ``Scalable multi-robot collaboration with large language models: Centralized or decentralized systems?'' \emph{arXiv preprint arXiv:2309.15943}, 2023.

\bibitem{lazaridou2016multi}
A.~Lazaridou, A.~Peysakhovich, and M.~Baroni, ``Multi-agent cooperation and the emergence of (natural) language,'' \emph{arXiv preprint arXiv:1612.07182}, 2016.

\bibitem{havrylov2017emergence}
S.~Havrylov and I.~Titov, ``Emergence of language with multi-agent games: Learning to communicate with sequences of symbols,'' \emph{Advances in neural information processing systems}, vol.~30, 2017.

\bibitem{lazaridou2018emergence}
A.~Lazaridou, K.~M. Hermann, K.~Tuyls, and S.~Clark, ``Emergence of linguistic communication from referential games with symbolic and pixel input,'' \emph{arXiv preprint arXiv:1804.03984}, 2018.

\bibitem{kottur2017natural}
S.~Kottur, J.~M. Moura, S.~Lee, and D.~Batra, ``Natural language does not emerge'naturally'in multi-agent dialog,'' \emph{arXiv preprint arXiv:1706.08502}, 2017.

\bibitem{lazaridou2020emergent}
A.~Lazaridou and M.~Baroni, ``Emergent multi-agent communication in the deep learning era,'' \emph{arXiv preprint arXiv:2006.02419}, 2020.

\bibitem{lazaridou2020multi}
A.~Lazaridou, A.~Potapenko, and O.~Tieleman, ``Multi-agent communication meets natural language: Synergies between functional and structural language learning,'' \emph{arXiv preprint arXiv:2005.07064}, 2020.

\bibitem{kim2024knowledge}
Y.~Kim, S.~Seo, J.~Park, M.~Bennis, S.-L. Kim, and J.~Choi, ``Knowledge distillation from language-oriented to emergent communication for multi-agent remote control,'' \emph{arXiv preprint arXiv:2401.12624}, 2024.

\bibitem{evtimova2017emergent}
K.~Evtimova, A.~Drozdov, D.~Kiela, and K.~Cho, ``Emergent communication in a multi-modal, multi-step referential game,'' \emph{arXiv preprint arXiv:1705.10369}, 2017.

\bibitem{mu2023ec2}
Y.~Mu, S.~Yao, M.~Ding, P.~Luo, and C.~Gan, ``Ec2: Emergent communication for embodied control,'' in \emph{Proceedings of the IEEE/CVF Conference on Computer Vision and Pattern Recognition}, 2023, pp. 6704--6714.

\bibitem{selcom1}
Z.~Ma, Y.~Luo, and J.~Pan, ``Learning selective communication for multi-agent path finding,'' \emph{IEEE Robotics and Automation Letters}, vol.~7, no.~2, pp. 1455--1462, 2022.

\bibitem{zhai2021decentralized}
Y.~Zhai, B.~Ding, X.~Liu, H.~Jia, Y.~Zhao, and J.~Luo, ``Decentralized multi-robot collision avoidance in complex scenarios with selective communication,'' \emph{IEEE Robotics and Automation Letters}, vol.~6, no.~4, pp. 8379--8386, 2021.

\bibitem{chen2023agentverse}
W.~Chen, Y.~Su, J.~Zuo, C.~Yang, C.~Yuan, C.~Qian, C.-M. Chan, Y.~Qin, Y.~Lu, R.~Xie \emph{et~al.}, ``Agentverse: Facilitating multi-agent collaboration and exploring emergent behaviors in agents,'' \emph{arXiv preprint arXiv:2308.10848}, vol.~2, no.~4, p.~6, 2023.

\bibitem{cascadingcode}
J.~Spracklen, R.~Wijewickrama, A.~Sakib, A.~Maiti, and M.~Jadliwala, ``We have a package for you! a comprehensive analysis of package hallucinations by code generating llms,'' \emph{arXiv preprint arXiv:2406.10279}, 2024.

\bibitem{deitke2022️}
M.~Deitke, E.~VanderBilt, A.~Herrasti, L.~Weihs, K.~Ehsani, J.~Salvador, W.~Han, E.~Kolve, A.~Kembhavi, and R.~Mottaghi, ``Procthor: Large-scale embodied ai using procedural generation,'' \emph{Advances in Neural Information Processing Systems}, vol.~35, pp. 5982--5994, 2022.

\bibitem{topviewvlm}
C.~Li, C.~Zhang, H.~Zhou, N.~Collier, A.~Korhonen, and I.~Vuli{\'c}, ``Topviewrs: Vision-language models as top-view spatial reasoners,'' \emph{arXiv preprint arXiv:2406.02537}, 2024.

\bibitem{liu2020when2com}
Y.-C. Liu, J.~Tian, N.~Glaser, and Z.~Kira, ``When2com: Multi-agent perception via communication graph grouping,'' in \emph{Proceedings of the IEEE/CVF Conference on computer vision and pattern recognition}, 2020, pp. 4106--4115.

\bibitem{arul2022dmca}
S.~H. Arul, A.~S. Bedi, and D.~Manocha, ``Dmca: Dense multi-agent navigation using attention and communication,'' \emph{arXiv preprint arXiv:2209.06415}, 2022.

\end{thebibliography}
}

\clearpage
\newpage

\appendix

\section{Prompts}
\label{sec:append:prompt}
Following are the prompts used for our experiments. The variables passed to the prompts are indicated by chevrons ($<\cdot>$).

\subsection{No Comm. - Ground}
\label{subsec:no_comm_ground}
\promptbox{System Prompt}{
I am a mobile ground robot equipped with a camera. You are a helper agent with access to the camera feed.\\
I am trying to reach an object in an indoor environment, and you should help me by giving me an action command to accomplish this task successfully.\\
You are only allowed to give me one of the following commands:\\
- MoveAhead: moves the robot in the forward direction by 0.25 meters.\\
- MoveBack: moves the robot in the back direction by 0.25 meters while looking forward.\\
- RotateRight: rotates the robot towards right by 90 degrees while staying at the same location.\\
- RotateLeft: rotates the robot towards left by 90 degrees while staying at the same location.\\
- DoNothing: the robot does not do anything and stays in place.\\
Give only an action command as the answer.
}

\promptbox{User Prompt}{
The target object is $<$target object$>$.\\
The last action was $<$last action$>$. The agent executed it $<$successfully/unsuccessfully$>$ and the resulting distance to the target object is $<$shortest path distance to the target.$>$ 
}

If the agent is unsuccessful in executing the action, the following is also added to the user prompt:\\
\promptbox{}{
The agent failed to execute it for the following reason: $<$blocking object$>$ is blocking the agent from moving.}

Additionally, the image from the ground agent's camera is provided.

\subsection{Communication}
\label{subsec:comm_dialogue}

\promptbox{Overhead Agent - System Prompt}{
You are an overhead agent with a top view in an environment. You must guide a ground robot agent to reach the $<$target object$>$.
}

\promptbox{Overhead Agent - User Prompt}{
Give me a dialogue to pass to the ground agent to reach the target. Respond with just the dialogue.\\
You must query about the environment $C_{len}$ times before giving any form of instruction.\\
\\
$<$Dialogue between the agents$>$
}

Additionally, the image from the overhead agent’s camera is provided.

\promptbox{Ground Agent - System Prompt}{
You are a ground agent with an egocentric view. You need to reach the $<$target object$>$.\\
You will be questioned by an overhead agent. Be very concise with your answers.
}

\promptbox{Ground Agent - User Prompt}{
$<$Dialogue between the agents$>$
}

Additionally, the image from the ground agent’s camera is provided.

\subsection{Execution}

\subsubsection{Cooperative Action}
\label{subsubsec:cooperative_action}

\promptbox{Overhead Agent - System Prompt}{
You are an Overhead Agent (OA) tasked with helping a Ground Agent (GA) equipped with a camera. You must guide GA to reach the $<$target object$>$ by prescribing an action command.
You will be given your conversation with the GA about the environment as well as the overhead image you currently see to make the decision. \\
Make sure to verify if you can see the $<$target object$>$ and if the conversation would lead it there. \\

The GA has not moved even if the conversation suggests so. Please look at the conversation and give one of the following commands from the GA's perspective: \\
- MoveAhead \\
- MoveBack \\
- RotateRight \\
- RotateLeft \\
- DoNothing \\

Give an action command as the answer.}

\promptbox{Overhead Agent - User Prompt}{
The target object is $<$target object$>$. Conversation: $<$Conversation between the agents$>$.
}

The ground agent follows the action provided by the overhead agent.

\subsubsection{Selective Action}
\label{subsubsec:selective_action}

First, we obtain a command from the overhead agent using the prompt in subsection~\ref{subsubsec:cooperative_action}. Then we ask the ground agent whether it wishes to follow the overhead agent's suggestion using the following prompt. 

\promptbox{Ground Agent - System Prompt}{
I am a mobile ground robot equipped with a camera. You are a helper agent with access to the camera feed.\\
I am trying to reach an object in an indoor environment. There is another agent with an overhead camera feed who is providing me with an instruction for this task.\\
Following are the valid action commands and their meaning for me:\\
- MoveAhead: moves the robot in the forward direction by 0.25 meters.\\
- MoveBack: moves the robot in the back direction by 0.25 meters while looking forward.\\
- RotateRight: rotates the robot towards right by 90 degrees while staying at the same location.\\
- RotateLeft: rotates the robot towards left by 90 degrees while staying at the same location.\\
- DoNothing: the robot does not do anything and stays in place.\\
I will tell you the target object and the action command that the overhead agent has given to me. You should tell me whether I should trust and follow the overhead agent's command (Yes) or whether I should disregard it and decide on my own (No).\\
You must answer in Yes or No.
}

\promptbox{Overhead Agent - System Prompt}{
The target object is  $<$target object$>$. The overhead agent prescribed the following action: $<$action command$>$.
}
Additionally, the image from the ground agent’s camera is provided.

If the ground agent decides not to follow the overhead agent's suggestion, we execute \textbf{No Comm.} scenario using the prompt given in subsection~\ref{subsec:no_comm_ground}.

\subsection{Preemptive Actions Classifier}
\label{subsec:append:preempt}
\promptbox{System Prompt}{
You will be given a conversation between an overhead agent and a ground agent. They are expected to talk about the environment around them and nothing else.\\
No actions should be assumed to have been taken during the period of the conversation.
Can you tell me if the conversation has any 'preemptive motions'?\\
For example, dialogues such as <'I have moved towards the room'> are 'preemptive motions', as they assume the agent has already executed a motion in the past.\\
If the motion is yet to be executed, and it is in the future, it is NOT a 'preemptive motion'.
Answer only with Yes or No.
}

\promptbox{User Prompt}{
<Conversation between the agents>
 }

\begin{figure}[h!]
    \centering
    \includegraphics[width=0.9\linewidth]{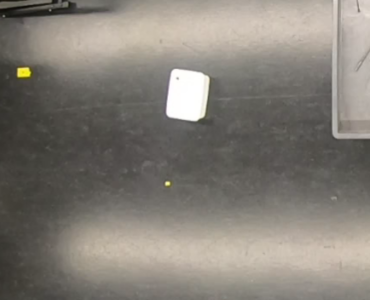}
    \caption{\textbf{Environment-Specfic Finetuning}: Upon passing the simulator prompts for conversation to this image, the overhead agent would misidentify the lights as a table. Explicitly mentioning the presence of white lights on the floor alleviates this issue.}
    \label{fig:reflection}
\end{figure}

\section{Adverserial Cases}
\label{sec:append:adverserial}
In our simulation experiments, we found that some observations may be perceived as adversarial or CAPTCHA-like inputs, resulting in error from \textit{GPT-4-turbo}.

\begin{figure}[h]
    \centering
    \begin{subfigure}{0.22\textwidth}
        \includegraphics[width=\linewidth]{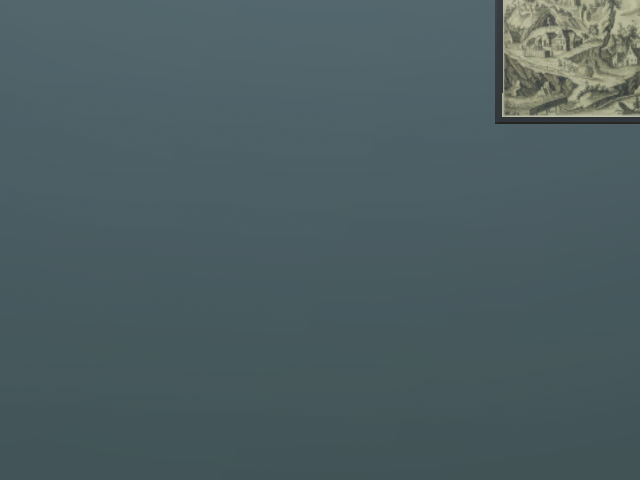}
        \caption{}
    \end{subfigure}\hfill
    \begin{subfigure}{0.22\textwidth}
        \includegraphics[width=\linewidth]{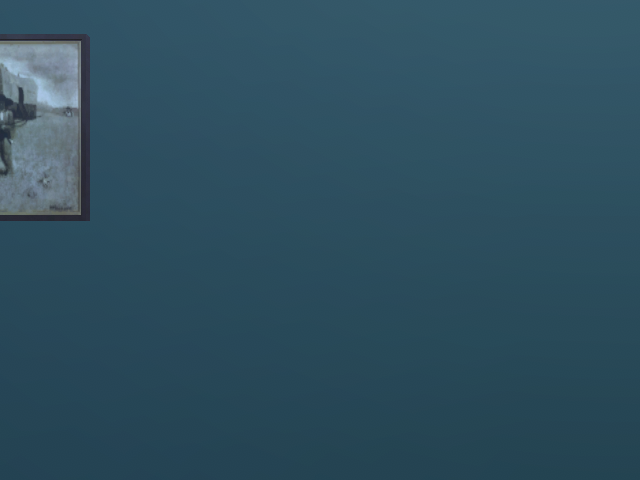}
        \caption{}
    \end{subfigure}\hfill
    \begin{subfigure}{0.22\textwidth}
        \includegraphics[width=\linewidth]{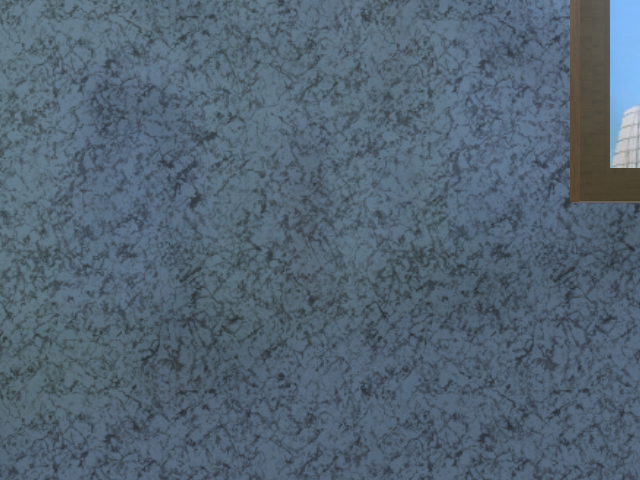}
        \caption{}
    \end{subfigure}\hfill
    \begin{subfigure}{0.22\textwidth}
        \includegraphics[width=\linewidth]{"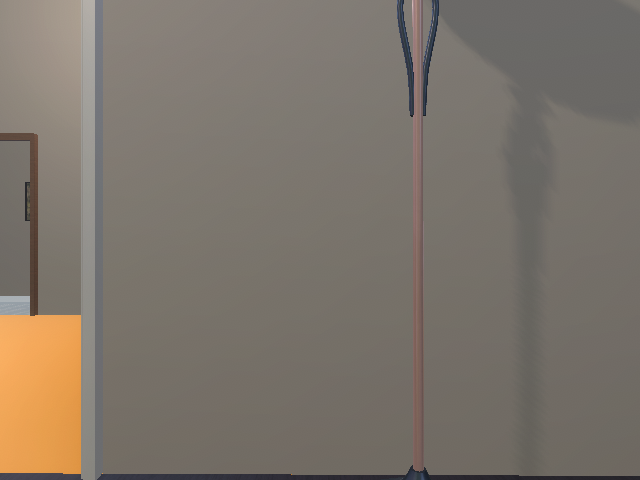"}
        \caption{}
    \end{subfigure}\hfill
    \begin{subfigure}{0.24\textwidth}
        \includegraphics[width=\linewidth]{"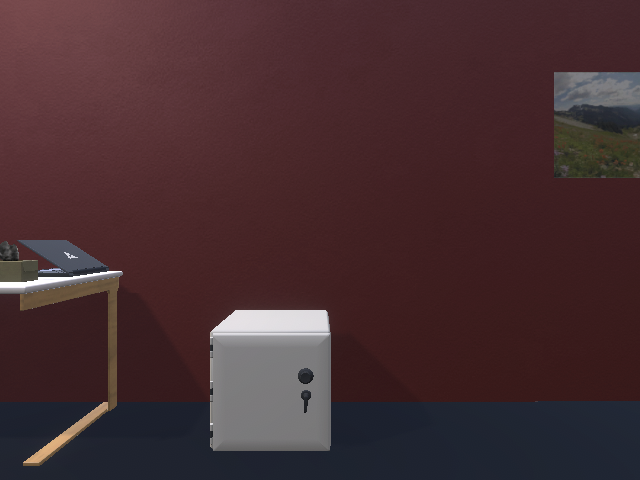"}
        \caption{}
        \label{fig:captcha}
    \end{subfigure}
    \caption{Agent observations which result in error with \textit{GPT-4-turbo}}
    \label{fig:adverserial}
\end{figure}

Fig.~\ref{fig:adverserial} shows some examples of such image inputs. As evident from these, images mainly containing uninteresting features such as walls cause the error. This could be due to safety measures put in place for \textit{GPT-4-turbo} to avoid adversarial attacks. Regardless, these cases showcase a vulnerability for robots equipped with similar LLMs. It is not uncommon for the robots to face walls during their navigation. However, observations from such trivial cases may require adding measures for embodied agents to deal with potential failures. We did not use the episodes with such examples in our evaluation. 

\begin{figure}[t!]
    \centering
    \includegraphics[width=\linewidth]{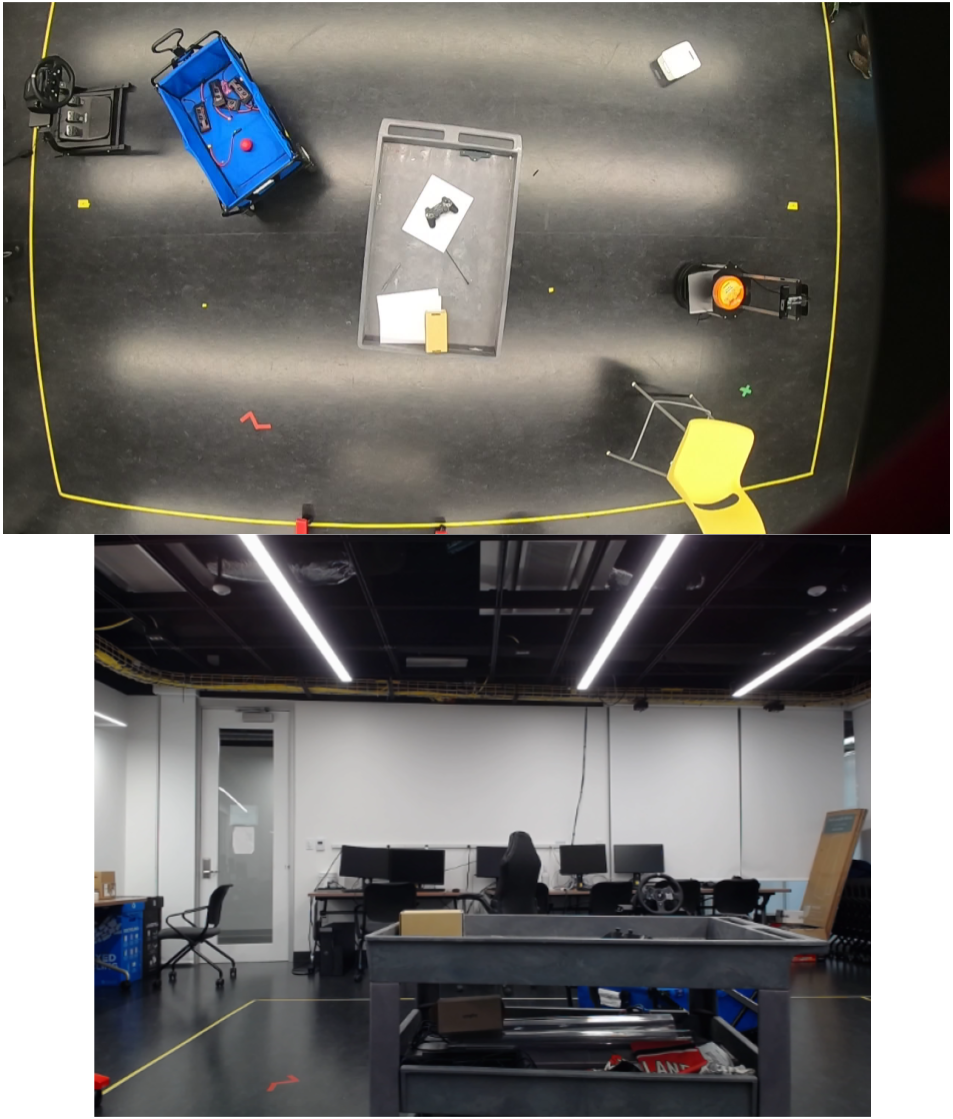}
    \caption{\textbf{LLaVA Communication Failiure}: The target here is the steering wheel. Notice that the agent is far away from it, and also that there are multiple objects in the way. The LLaVA conversation fails to capture the objects, and also fails to localize the ground agent's location. Instead, it mentions that there are no objects or features blocking it's path to the steering wheel. This hallucination leads to poor action prediction.}
    \label{fig:llava}
\end{figure}

\section{Real World Experiments}
\subsection{Finetuning Prompts}

Our prompt in the simulator is described in Appendix \ref{sec:append:prompt}. In the real world, we add additional constraints to allow the agent to localize the agent with the targets detected.

\promptbox{Overhead Agent - System Prompt}{
You are an Overhead Agent (OA) with a top view in an environment. You must identify and guide a Ground Agent (GA) to reach the <target object>. \\
\textcolor{magenta}{The GA is a turtlebot with a circular orange box on it. Be sure to use the ground agent's perspective. Also, note there is a reflection from multiple rectangular white lights on the black floor.}
}

\promptbox{Overhead Agent - User Prompt}{
\textcolor{magenta}{Give me a dialogue questioning the GA about its surroundings. The GA can not move while speaking with you. \\
You must query the GA about the environment and try to estimate its position and orientation with respect to the <target object>}. You have {$C_{Len}$ - 1} questions left to ask. \\
$<$Dialogue between the agents$>$
}

\promptbox{Ground Agent - System Prompt}{
You are a Ground Agent with an egocentric view. You need to reach the $<$target object$>$.\\
You will be questioned by an Overhead Agent about things you see. Be very concise with your answers.
}

\promptbox{Ground Agent - User Prompt}{
$<$Dialogue between the agents$>$
}

Note here that we ask the OA to use environmental cues to figure out the position and orientation of the agent in the communication phase. This explicit mention helps curb preemptive hallucinations. Additionally, we also provide unique identifying details about the ground agent (with an orange box here, for instance).

Additional information such as \textit{``note there is a reflection from multiple rectangular white lights on the black floor''} is environment-specific, and allows the VLM to adapt better. Prior to providing such an instruction, the agent kept hallucinating the white reflections on the floor to be a table. This is illustrated in Fig. \ref{fig:reflection}.

\subsection{VLM Ablations}

Apart from GPT-4V, we also conducted experiments with InstructBLIP with FLAN-T5 and Vicuna-7B bases, as well as LLaVA-1.5 finetuned on LORA. However, we observe that the conversations with these models suffers from significant hallucination in the form of repetition and false information. We combine the system and user prompts presented earlier and pass this as a single prompt here.
For instance, consider the example below:
\promptbox{Conversation with InstructBLIP-FlanT5-XXL}{
\begin{quote}
    OA: Ground Agent, What do you see in front of you? \\
    GA: There is a table and chairs. \\
    OA: What color is the table? \\
    GA: It is blue. \\
    OA: What color is the chair? \\
    GA: It is red. \\
    OA: What color is the carpet? \\
    GA: It is black. \\
    OA: What color is the wall? \\
    GA: It is white. \\
    OA: What color is the door? \\
    GA: It is blue. \\
    OA: What color is the window? \\
    GA: It is green. \\
    OA: What color is the sky? \\
    GA: It is blue. \\
    OA: What color is the grass? \\
    GA: It is green. \\
    OA: What color is the water? \\
    GA: It is blue. \\
\end{quote}
}

Notice the conversation starts off with an expected query asking about what was in front of the agent, but quickly derails into dialogue that is not useful in guiding the ground agent. The questions also become more vague towards the end, and are not grounded in the environment anymore.

\promptbox{Conversation with InstructBLIP-Vicuna-7B}{
\begin{quote}
    OA: Can you describe what you see around you? \\
    GA: I'm sorry, but I'm a Ground Agent with a limited view of the area. I can't see what's going on up there. \\
    OA: I'm sorry, but I'm an Overhead Agent with a bird's-eye view of the area. I'm looking for a Ground Agent to guide me to the steering wheel.
\end{quote}
}

Despite Vicuna-7B being finetuned on conversations/dialogue data, notice the apparent confusion that occurs here. The overhead agent assumes that the ground agent is going to guide it to the steering wheel, while the opposite is being explicitly stated to the OA, i.e. "You must
identify and guide a Ground Agent (GA) to reach the $<$target object$>$."

\promptbox{Conversation with LLaVA-1.5-13B}{
\begin{quote}
    OA: Hello GA, can you tell me where you are in relation to the steering wheel? Are you close to it or far away? Also, can you see any other objects or obstacles in your path that might be blocking your way to the steering wheel? \\
    GA: I am in a room with a table and a chair. The steering wheel is located in the room, but I am not close to it. There are no obstacles in my path, but I need to navigate through the room to reach the steering wheel. \\
    OA: Great, GA. Now, can you tell me the direction you need to take to reach the steering wheel? Is it to your left, right, or straight ahead? Also, can you see any other objects or features in the room that might help you navigate towards the steering wheel? \\
    GA: \textcolor{red}{I need to take a right turn to reach the steering wheel. In the room, there is a table and a chair, which might help me navigate towards the steering wheel.} \\
    OA: Thank you for the information, GA. Now, can you tell me if there are any other objects or features in the room that might be blocking your path or making it difficult to reach the steering wheel? Also, can you see any other objects or obstacles in the room that might be helpful in guiding you towards the steering wheel? \\
    GA: \textcolor{red}{There are no objects or features in the room that are blocking my path} or making it difficult to reach the steering wheel. The table and chair in the room might be helpful in guiding me towards the steering wheel.
\end{quote}
}

In the LLaVA conversation, notice that the GA assues the OA that there are no objects in its path to reach the steering wheel. It also mentions a table and chair. However, upon seeing the actual OA and GA images (Figure \ref{fig:llava}), we note that neither the steering wheel is clearly ahead, nor is there only a table or chair in the room. Moreover, the OA does not even attempt to utilize it's own understanding of the environment to identify and guide the GA as it was asked to.

\subsection{Localization Failures}
\label{sec:append:localization_fail}
In a communication setup, one may expect the cross-view localization to be trivial as VLM can match the objects in the two images (from GA and OA) to guess the location of the robot. However, we found that VLM struggles at doing so. 

\begin{figure}[h]
    \centering
    \begin{subfigure}{0.49\textwidth}
        \includegraphics[width=\linewidth]{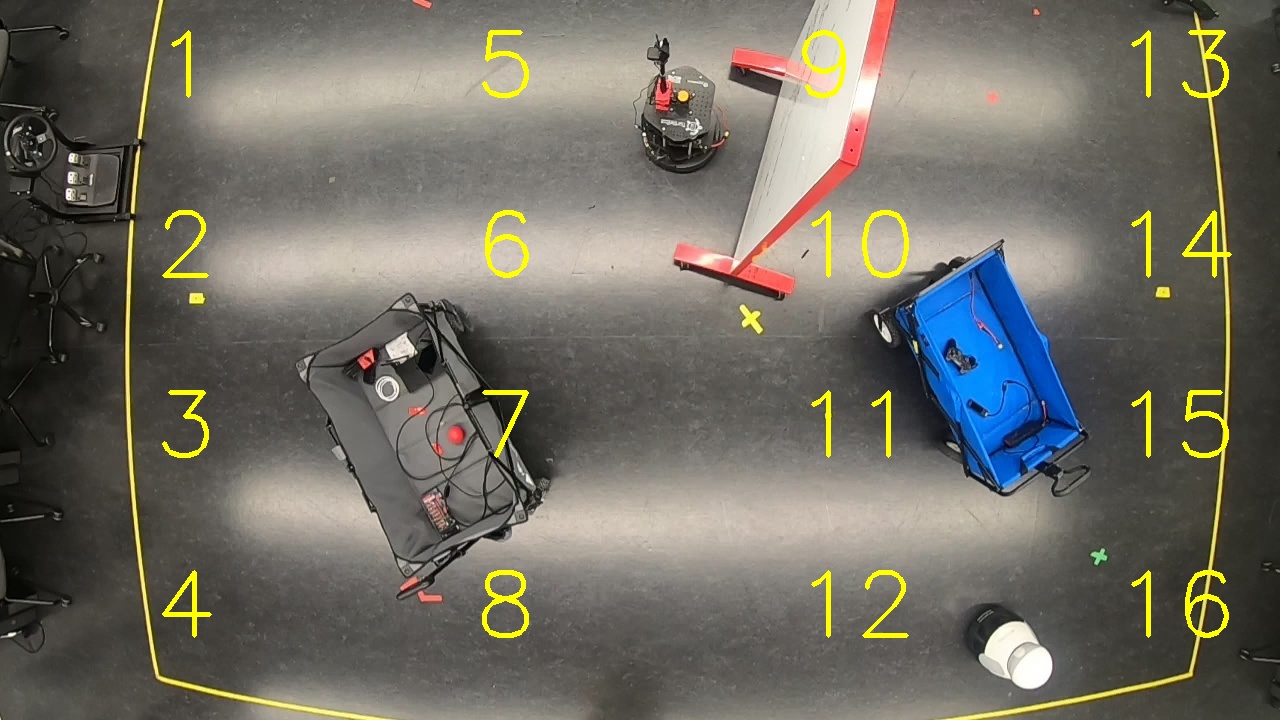}
        \caption{Predicted position: 6}
    \end{subfigure}\hfill
    \begin{subfigure}{0.49\textwidth}
        \includegraphics[width=\linewidth]{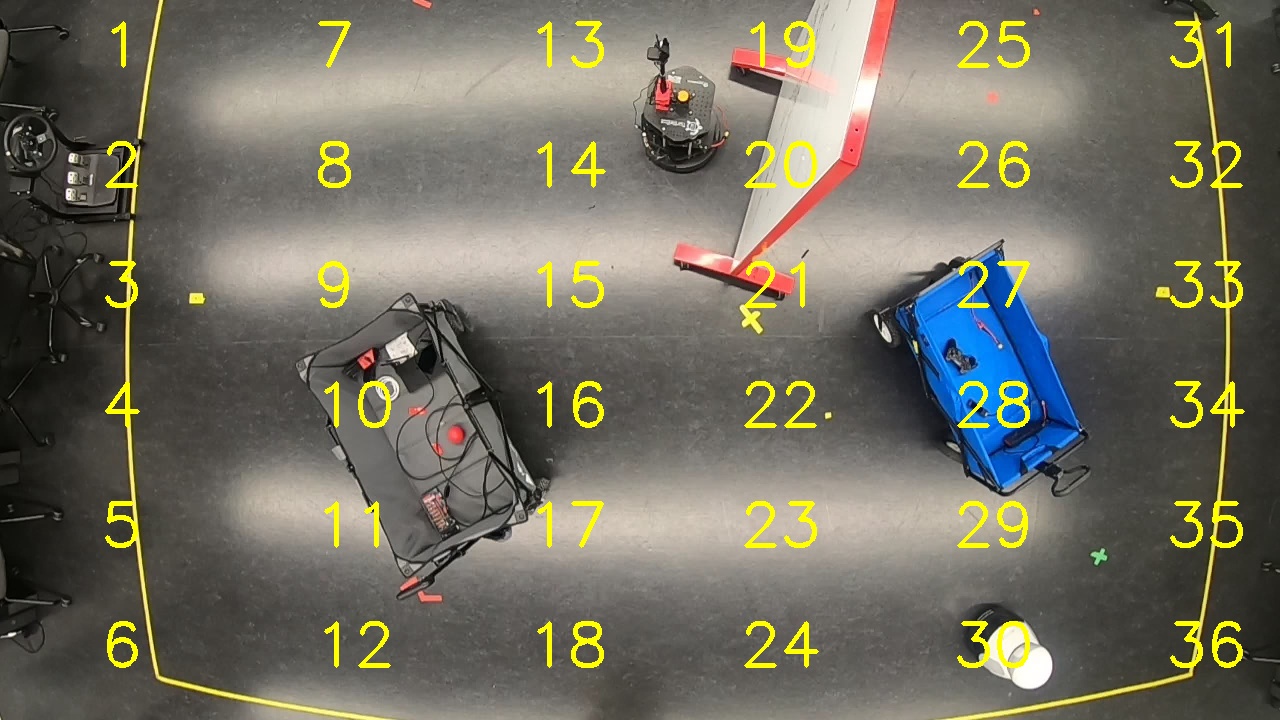}
        \caption{Predicted position: 16}
    \end{subfigure}\vfill
    \begin{subfigure}{0.49\textwidth}
        \includegraphics[width=\linewidth]{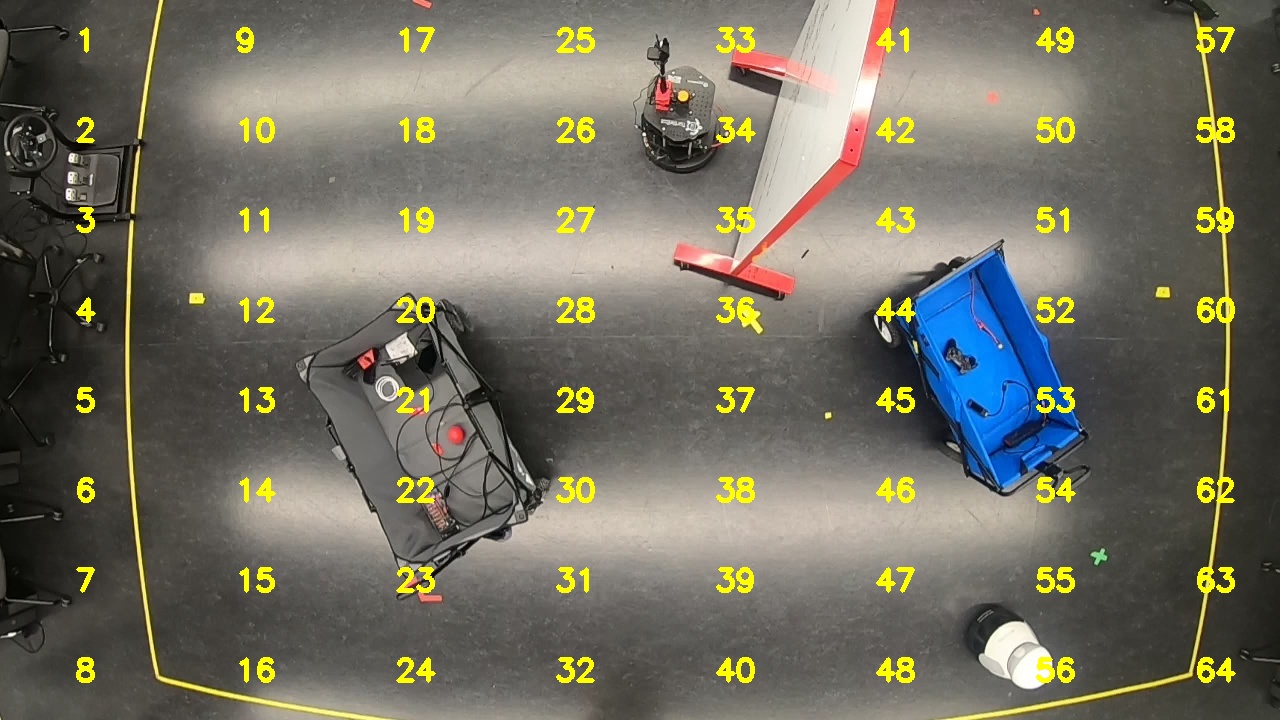}
        \caption{Predicted position: 21}
    \end{subfigure}\hfill
    \begin{subfigure}{0.49\textwidth}
        \includegraphics[width=\linewidth]{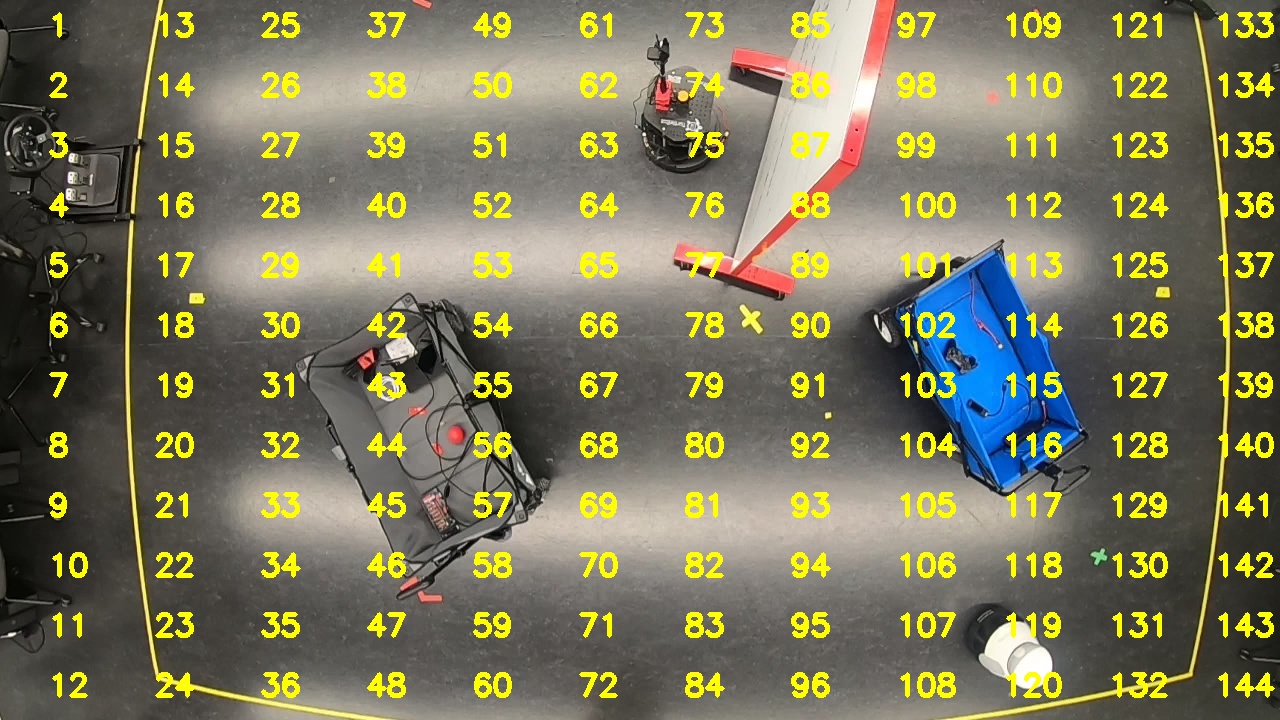}
        \caption{Predicted position: 42}
    \end{subfigure}\hfill
    \caption{Localization response from \textit{GPT-4-turbo} for varying extent of labelling}
    \label{fig:localization_size}
\end{figure}

\begin{figure}[h]
    \centering
    \begin{subfigure}{0.49\textwidth}
        \includegraphics[width=\linewidth]{figs/image_23_grid_6.jpg}
        \caption{Predicted position: 16}
    \end{subfigure}\hfill
    \begin{subfigure}{0.49\textwidth}
        \includegraphics[width=\linewidth]{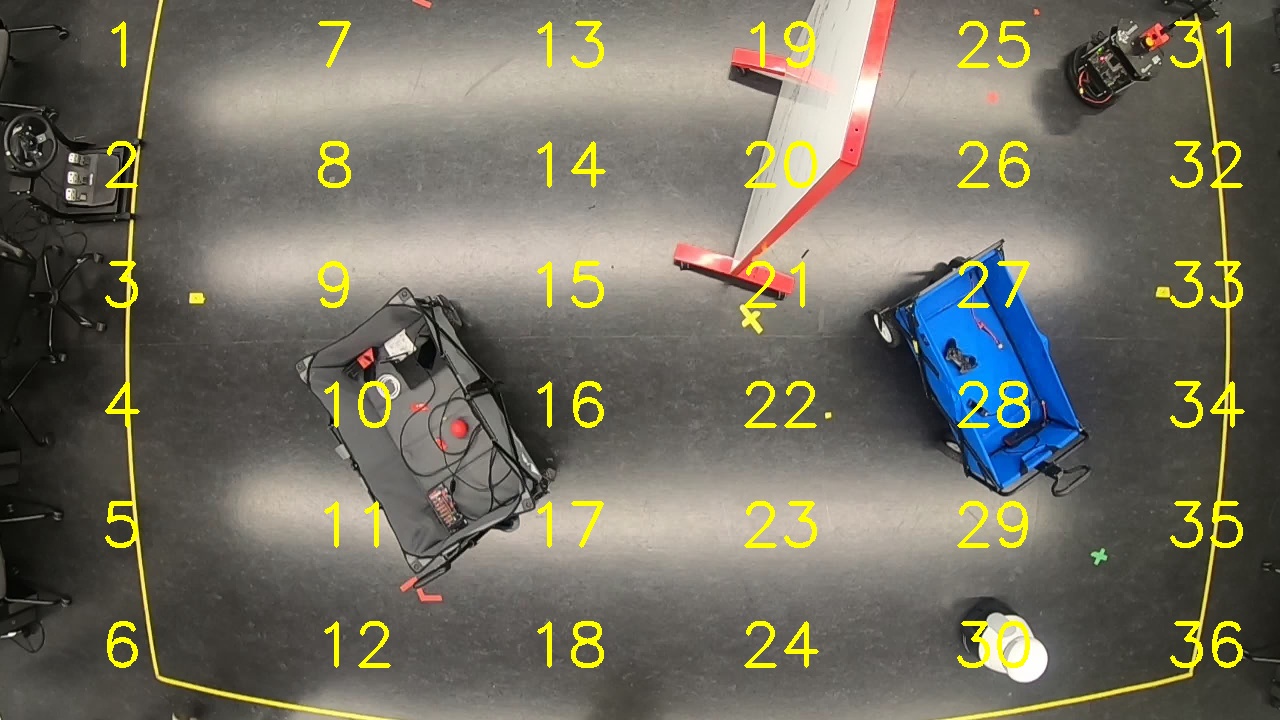}
        \caption{Predicted position: 30}
    \end{subfigure}\vfill
    \begin{subfigure}{0.49\textwidth}
        \includegraphics[width=\linewidth]{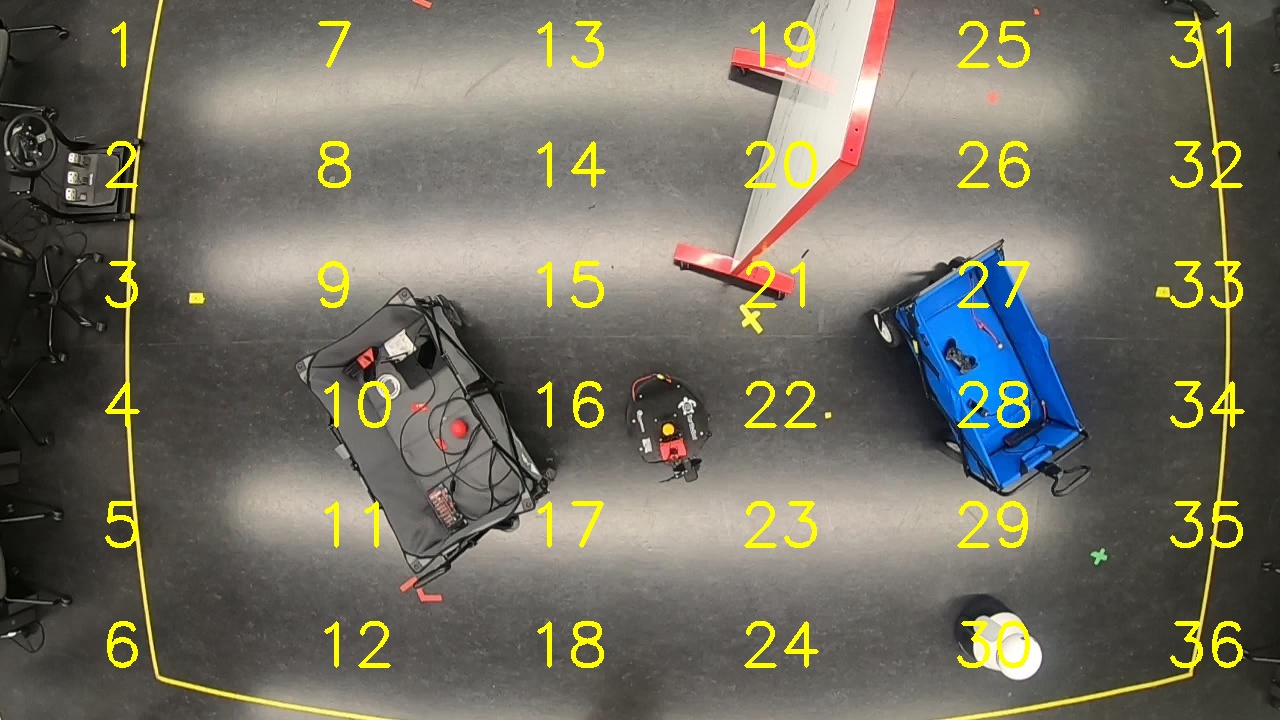}
        \caption{Predicted position: 16}
    \end{subfigure}\hfill
    \begin{subfigure}{0.49\textwidth}
        \includegraphics[width=\linewidth]{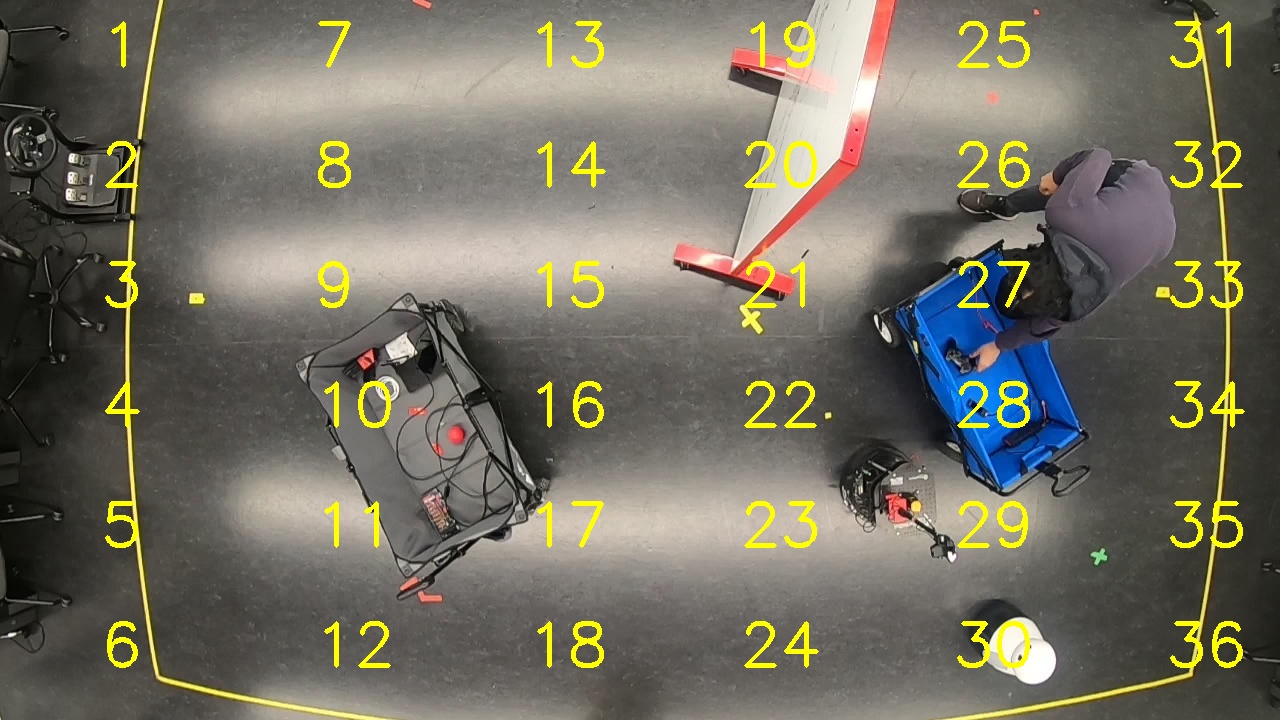}
        \caption{Predicted position: 9}
    \end{subfigure}\hfill
    \caption{Localization response from \textit{GPT-4-turbo} for different locations of the Turtlebot2}
    \label{fig:localization_pose}
\end{figure}

For this experiment, we use a Turtlebot2 robot in an indoor environment with a few other objects (carts with objects inside, whiteboard, driving simulator, etc). The Turtlebot2 is black in color and has a range sensor with an orange cap on top.  For the overhead view, we put a GoPro 7 camera on top. Inspired from PIVOT~\cite{nasiriany2024pivot}, we added labels to the overhead image and asked the VLM agent to find `a black Turtlebo2 robot with an orange sensor on top'. To add the labels, we split the image into grids with 16, 36, 64, and 144 cells and varied the label sizes to ensure that the labels were equally spaced without any overlap. As shown in Fig.~\ref{fig:localization_size}, the results are comparatively better with smaller grids but not none of the results are precise.
A closer look reveals that the VLM may be confused due to similar-looking objects in the scene (a grey cart with red and range objects inside). This is also shown in Fig.~\ref{fig:localization_pose}, which shows the Turtlebot at different locations and the position predicted by the VLM. 

\begin{figure}[h]
    \centering
    \begin{subfigure}{0.4\textwidth}
        \includegraphics[width=\linewidth]{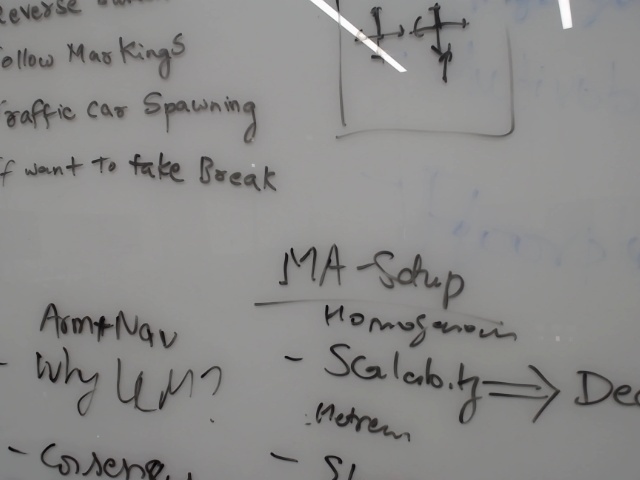}
        \caption{Predicted position: 16}
    \end{subfigure}\hfill
    \begin{subfigure}{0.4\textwidth}
        \includegraphics[width=\linewidth]{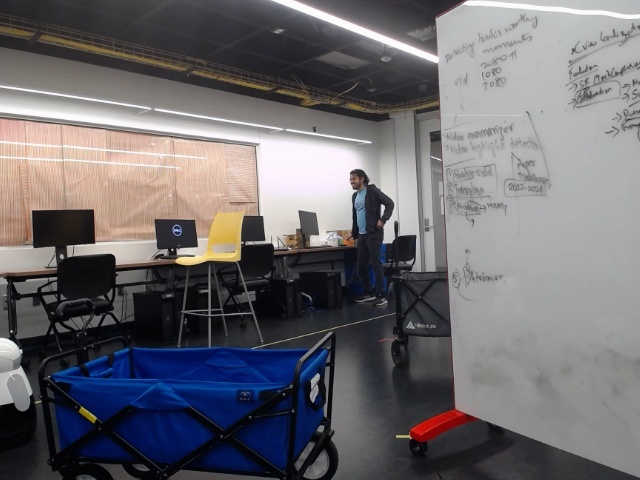}
        \caption{Predicted position: 10}
    \end{subfigure}\vfill
    \begin{subfigure}{0.4\textwidth}
        \includegraphics[width=\linewidth]{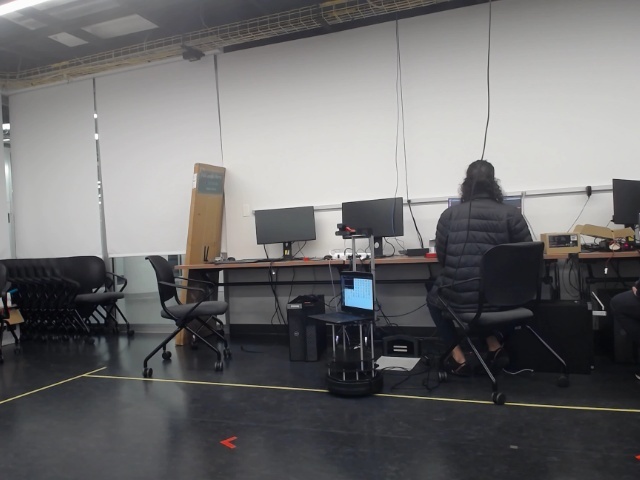}
        \caption{Predicted position: 16}
    \end{subfigure}\hfill
    \begin{subfigure}{0.4\textwidth}
        \includegraphics[width=\linewidth]{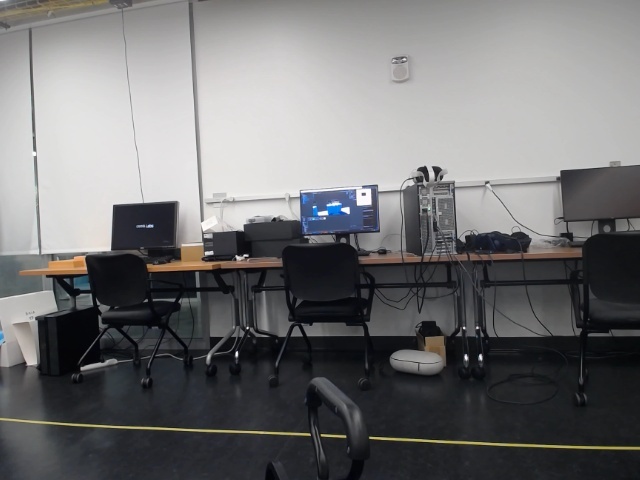}
        \caption{Predicted position: 10}
    \end{subfigure}\hfill
    \caption{Egocentric images from the camera on Turtlebot2. These images correspond to images of robot locations shown in  Fig.~\ref{fig:localization_pose}}
    \label{fig:localization_ego}
\end{figure}

We also tried adding the egocentric images from the camera on Turtlebot, as shown in Fig.~\ref{fig:localization_ego}, as additional inputs to VLM to help with cross-view localization, but they do not improve the results. 

\subsection{Identifying Correct Action from Dialogue}
\begin{figure}[h]
    \centering
    \includegraphics[width=\linewidth]{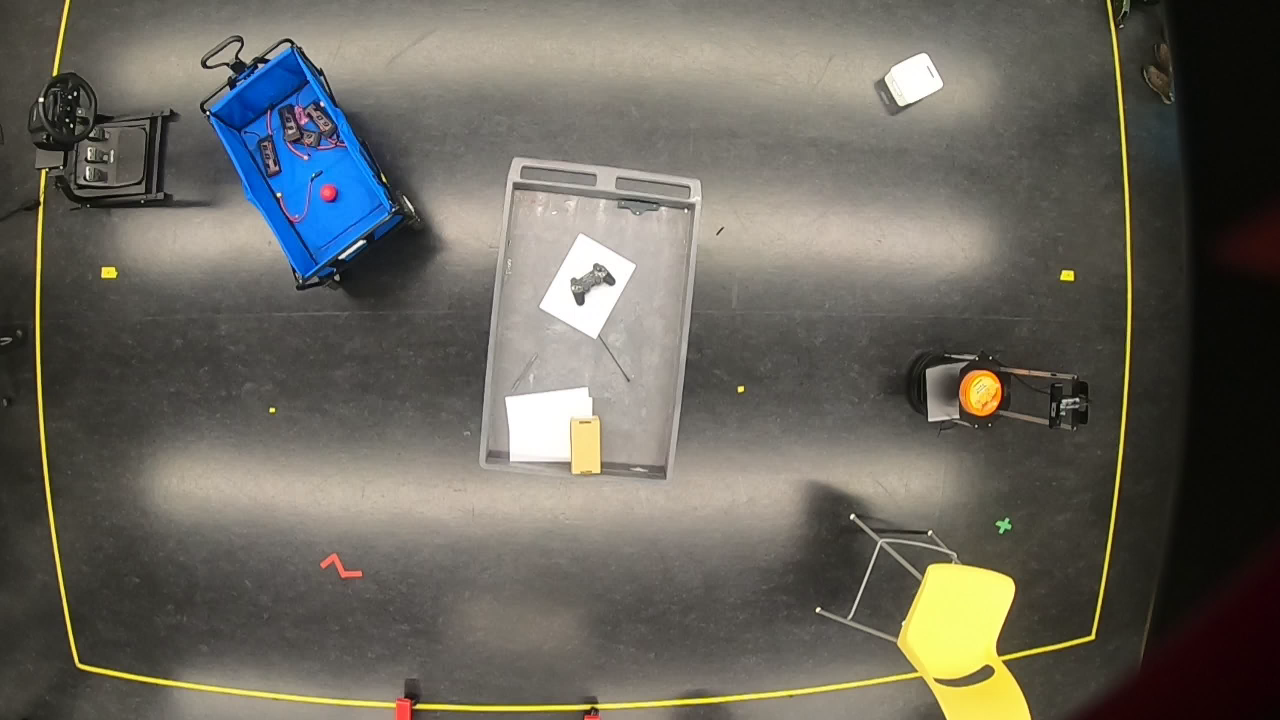}
    \caption{\textbf{Corrective Action with GA}: In this case, the agent decides to rotate by 180 degrees first and then move towards the left. The OA in the execution phase correctly suggests that the same can be accomplished by rotating towards right.}
    \label{fig:180deg_case}
\end{figure}

In many cases, the planning failure occurs due to the preemptive action hallucination in the conversation. During the execution phase, $\mathcal{E}$, the OA looks at this conversation and tries to execute an action to continue toward the target object. However, it may provide the right action if prompt fine-tuning reduces the preemptive action hallucination. Fig.~\ref{fig:180deg_case} shows this case where the target is the white basket on the ground and the GA is currently looking towards the center of the image.

\textbf{Conversation Phase:}\\
\textbf{OA:} Can you describe any nearby objects or structures that are directly in front of you?\\
\textbf{GA:} Directly in front of me, there is a work area with several desks and office chairs. There's a person sitting at one of the desks, working on a computer. To the left, there's a door, and overhead are strip lighting fixtures and exposed ceiling infrastructure. The foreground shows a metal cart with various items, possibly tools or mechanical components.
\textbf{OA:} What do you see to your right and behind you?\\
\textbf{GA:} To my right, there is an office area with multiple desks equipped with computer monitors and an office chair. Behind me, there is more of the workspace visible, including additional desks and chairs. A whiteboard is visible on the right wall, and there's a large technical rigging above the workspace, likely part of the lab or office's infrastructure.\\
\textbf{OA:} Based on the information you provided and the overhead view, you are currently facing away from the small white basket on the floor. To reach the basket, first rotate 180 degrees, then move forward until you reach the basket positioned on your left side after the turn.\\
\textbf{GA:} Understood. I will rotate 180 degrees and then advance to locate the small white basket positioned to the left after the turn.\\

\textbf{Execution Phase:}\\
\textbf{Prescribed Action:} RotateRight

When asked to provide a reason, GPT-4 responded with the following:\\
\textbf{OA:} Based on the overhead image and the Ground Agent's plan to rotate 180 degrees, a single command of "Rotate Right" will initiate this maneuver. 

We also note that the GA described the objects behind it even though it can only get observations from the front-looking camera. OA was able to correctly estimate the GA's location relative to the white basket as it gathered information about its observations to correctly estimate its pose. 

\section{Selective Communication}
 While communication with the OA helps the GA with ObjectNav, bandwidth, and cost restrictions may require a selective approach toward initiating communication~\cite{liu2020when2com, arul2022dmca}. To investigate such situations, we developed a variation of \textit{Cooperative Action} where the ground agent is first asked if it wishes to seek help from the overhead agent. The view from the ground agent camera is used as an additional input here. We use the simulation setup as described in section~\ref{subsec:simualtion_setup} for 100 episodes and report our findings in Table~\ref{tab:selective_comm}. We observe that such a \textit{selective communication} setup results in an improvement over the Cooperative Action setup, suggesting a judicious assistance-seeking behavior. However, we note that it comes at the cost of additional prompting at every episode.
 
\begin{table}[ht]
\centering
\begin{tabular}{ccrrr}
\toprule
\textbf{Comm. Length ($\mathcal{C}_{Len}$)} & \textbf{OSR \%} $\uparrow$ & \textbf{SPL \%} $\uparrow$ \\
\midrule
1 & 20.00 & 19.20 \\
3 & 23.00 & 21.65 \\
5 & 26.00 & 24.49 \\
\bottomrule
\end{tabular}
\vspace{0.3cm}
\caption{\textbf{Selective Communication Performance}: We find that letting the GA choose whether to communicate with the agent or not results in a better ObjectNav performance than \textbf{Cooperative Action}, but requires additional prompting at each step.}
\label{tab:selective_comm}
\end{table}

\section{Communication Wordclouds}
\begin{figure}[ht]
    \centering
    \begin{subfigure}{0.49\textwidth}
        \includegraphics[width=\linewidth]{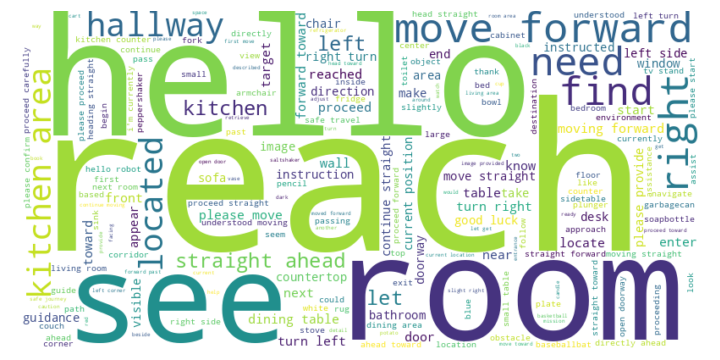}
        \caption{Word cloud for $\mathcal{C}_{Len} = 1$}
    \end{subfigure}\hfill
    \begin{subfigure}{0.49\textwidth}
        \includegraphics[width=\linewidth]{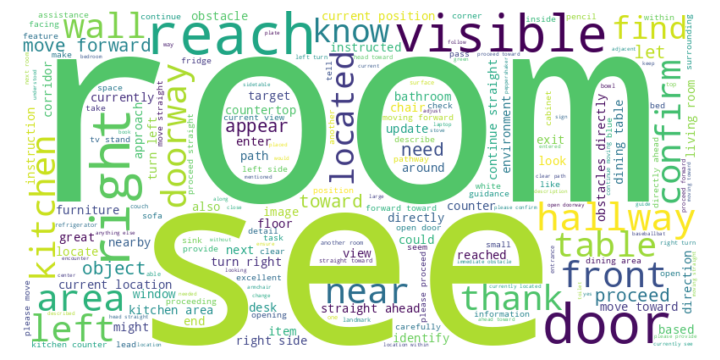}
        \caption{Word cloud for $\mathcal{C}_{Len} = 3$}
    \end{subfigure}\vfill
    \begin{subfigure}{0.49\textwidth}
        \includegraphics[width=\linewidth]{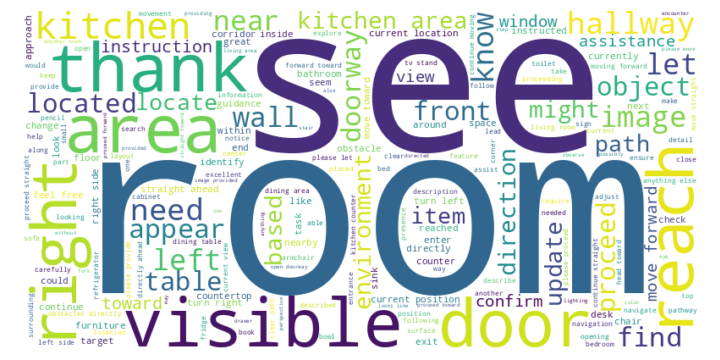}
        \caption{Word cloud for $\mathcal{C}_{Len} = 5$}
    \end{subfigure}\hfill
    \caption{World clouds over the communication between the GA and OA in \textit{Selective Action} setup. Communication in \textit{Cooperative Action} setup also exhibits similar patterns.}
    \label{fig:wordclouds}
\end{figure}

To identify the patterns in GC, we look at the word cloud over the communication between GA and OA. Fig.~\ref{fig:wordclouds} shows the word clouds for the \textit{Selective Action} setup. We observe that across all  $\mathcal{C}_{Len}$, the `rooms', `see', and `visible' are prominent as the OA always asks GA questions about surroundings and whether it sees the target object. 

For $\mathcal{C}_{Len} = 1$, we also see more prominence of action commands such as `forward' and `move' as the limited communication length may make the OA may suggest actions to reach the target right away. Some of these suggestions may instruct the GA to `reach' certain objects before reaching the target object. We also see `hello' having more weight for this case compared to other communication lengths as the OA greets the GA at the beginning of the conversation in most episodes. 

$\mathcal{C}_{Len} = 3$ have similar word clouds for words related to queries about the surroundings. Due to the large communication length, the OA may have the chance to suggest going to other rooms through `doors'. We observe similar trends for the \textit{Cooperative Action} setup as well.

\end{document}